\documentclass[letterpaper]{article} % DO NOT CHANGE THIS
\usepackage[submission]{aaai25}  % DO NOT CHANGE THIS
\usepackage{times}  % DO NOT CHANGE THIS
\usepackage{helvet}  % DO NOT CHANGE THIS
\usepackage{courier}  % DO NOT CHANGE THIS
\usepackage[hyphens]{url}  % DO NOT CHANGE THIS
\usepackage{graphicx} % DO NOT CHANGE THIS
\usepackage{natbib}  % DO NOT CHANGE THIS AND DO NOT ADD ANY OPTIONS TO IT
\usepackage{caption} % DO NOT CHANGE THIS AND DO NOT ADD ANY OPTIONS TO IT
\usepackage{algorithm}
\usepackage{algorithmic}

\usepackage{newfloat}
\usepackage{listings}
\DeclareCaptionStyle{ruled}{labelfont=normalfont,labelsep=colon,strut=off} % DO NOT CHANGE THIS
\floatstyle{ruled}
\newfloat{listing}{tb}{lst}{}
\floatname{listing}{Listing}

\usepackage{amsmath}
\usepackage{multirow}
\usepackage{booktabs}
\usepackage{subcaption}
\usepackage{amssymb}
\usepackage[table]{xcolor}
\usepackage{threeparttable}
\usepackage{longtable}
\usepackage[most]{tcolorbox}
\newtcolorbox{prompt}[1]{
    enhanced,
    colback=gray!20,
    colframe=black,
    boxrule=0.3pt,
    arc=3mm,
    left=2pt,
    right=2pt,
    boxsep=3pt,
    fonttitle=\small\bfseries,
    title=#1,
    fontupper=\scriptsize
}

\title{Doc2SAR: A Synergistic Framework for High-Fidelity Extraction of Structure-Activity Relationships from Scientific Documents}

\author {
    Jiaxi Zhuang\equalcontrib,
    Kangning Li\equalcontrib,
    Jue Hou, 
    Mingjun Xu, 
    Zhifeng Gao,
    Hengxing Cai\thanks{Corresponding authors.}
}
\affiliations {
    DP Technology, Beijing, China\\
}
\begin{document}

\maketitle

\begin{abstract}
Extracting molecular structure-activity relationships (SARs) from scientific literature and patents is essential for drug discovery and materials research. However, this task remains challenging due to heterogeneous document formats and limitations of existing methods. Specifically, rule-based approaches relying on rigid templates fail to generalize across diverse document layouts, while general-purpose multimodal large language models (MLLMs) lack sufficient accuracy and reliability for specialized tasks, such as layout detection and optical chemical structure recognition (OCSR). To address these challenges, we introduce \textbf{DocSAR-200}, a rigorously annotated benchmark of 200 scientific documents designed specifically for evaluating SAR extraction methods. Additionally, we propose \textbf{Doc2SAR}, a novel synergistic framework that integrates domain-specific tools with MLLMs enhanced via supervised fine-tuning (SFT). Extensive experiments demonstrate that Doc2SAR achieves state-of-the-art performance across various document types, significantly outperforming leading end-to-end baselines. Specifically, Doc2SAR attains an overall Table Recall of \textbf{80.78\%} on DocSAR-200, exceeding end2end GPT-4o by \textbf{51.48\%}. Furthermore, Doc2SAR demonstrates practical usability through efficient inference and is accompanied by a web app. The code and data are provided in the supplementary materials.
\end{abstract}

\section{Introduction}
\label{sec:introduction}

The extraction of molecular structure-activity relationships (SARs) from scientific literature and patents is fundamental for accelerating the discovery of novel drugs and advanced materials \cite{sar0,sar1,sar2}. Accurate and efficient extraction of molecular activity data not only supports computational modeling but also significantly reduces innovation cycles in pharmaceutical and materials science research \cite{drews2000drug,dara2022machine,deng2022artificial,sadybekov2023computational}. Despite its critical importance, automated SAR extraction remains challenging due to the complexity and heterogeneity of scientific documents.

\begin{figure*}
    \centering
    \includegraphics[width=\linewidth]{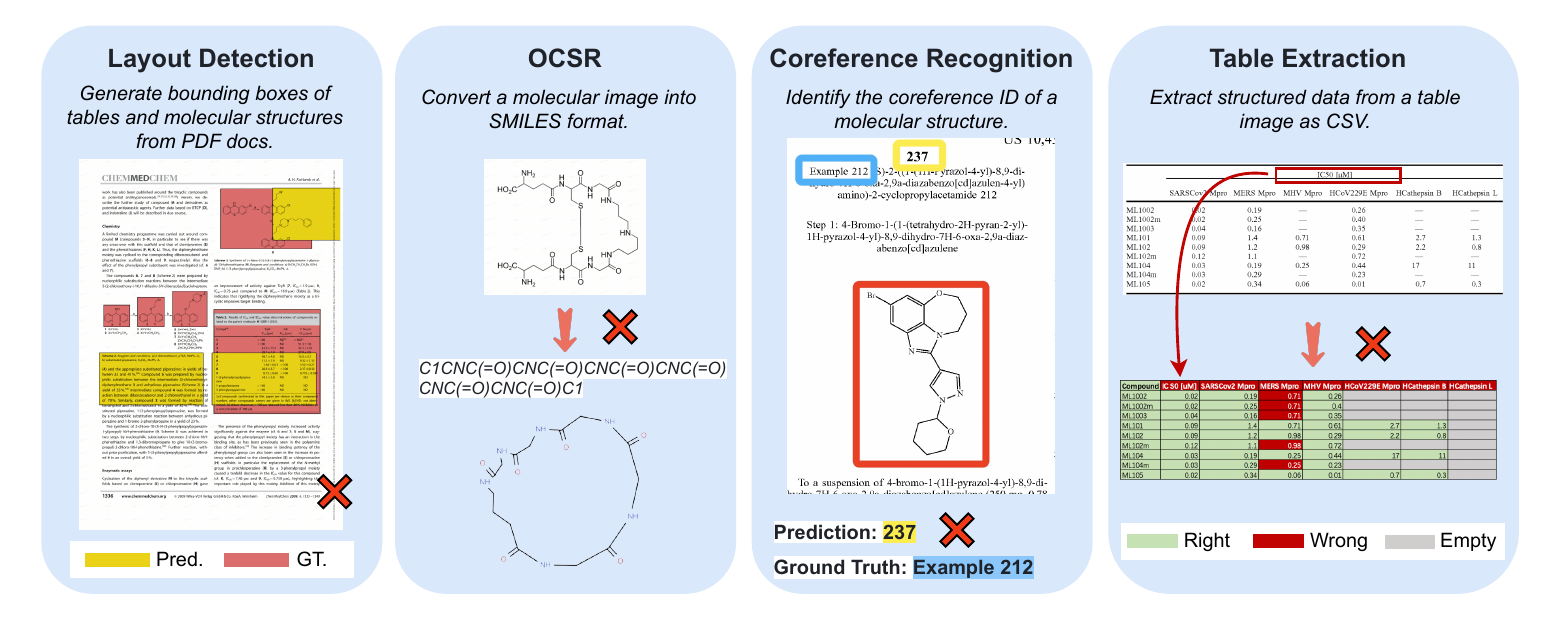}
    \caption{Motivation for a modular framework: An end-to-end GPT-4o model shows compounding errors across crucial stages of SAR extraction.}
    \label{fig:failure-gpt4o}
\end{figure*}

Two major issues currently impede progress in automated SAR extraction. Firstly, there is a notable \textbf{lack of standardized benchmark datasets} specifically designed to evaluate and advance methods for this complex, multimodal information extraction task. Secondly, existing automated extraction methods exhibit critical deficiencies: traditional rule-based methods employing predefined templates are inflexible and fail to generalize across \textbf{diverse document formats} \cite{chemex,chemical,chemdataextractor}; while general-purpose multimodal large language models (MLLMs), though promising \cite{llm4ex,llm4ex2}, still \textbf{fall short in specialized subtasks} \cite{sciassess,sciriff,galactica,scibert}. As our failure analysis in Figure~\ref{fig:failure-gpt4o} demonstrates, the practical utility of monolithic, end-to-end GPT-4o \cite{gpt4o} is limited by a cascade of compounding errors. This inherent brittleness highlights the need for a more sophisticated, modular approach.

To address these challenges, we make two primary contributions. First, we introduce DocSAR-200, a meticulously curated benchmark of 200 diverse scientific documents designed to provide a standardized testbed for evaluating SAR extraction methods. Second, to tackle the complexities presented by this benchmark, we propose Doc2SAR, a novel synergistic framework. This framework integrates domain-specific tools like optical chemical structure recognition (OCSR) and fine-tuned MLLMs to precisely solve complex sub-tasks, with robust rule-based methods for post-processing.
Our key contributions can be summarized as:
\begin{enumerate}
    \item We introduce DocSAR-200, a comprehensive and rigorously annotated benchmark specifically tailored for evaluating progress in automated SAR extraction.
    \item We propose Doc2SAR, a novel synergistic framework that intelligently combines domain-specific tools and MLLMs enhanced via supervised fine-tuning (SFT) to precisely solve complex sub-tasks, demonstrating a superior approach to naive end-to-end models.
    \item Extensive experiments validate the effectiveness of Doc2SAR, which achieves state-of-the-art performance by attaining an overall Table Recall of 80.78\% on the DocSAR-200 benchmark, representing an improvement of 51.48\% over the strongest end-to-end baseline.
    % \item Doc2SAR demonstrates strong practical applicability, efficiently processing over 100 PDFs per hour on a single RTX 4090 GPU. Additionally, it features an interactive web interface for easy visualization, verification, and manual refinement.
    \item Doc2SAR demonstrates strong practical applicability, efficiently processing over 100 PDFs per hour on a single RTX 4090 GPU. Additionally, it features an interactive web interface for visualization and manual refinement.
\end{enumerate}

\begin{figure}
    \centering
    \includegraphics[width=\linewidth]{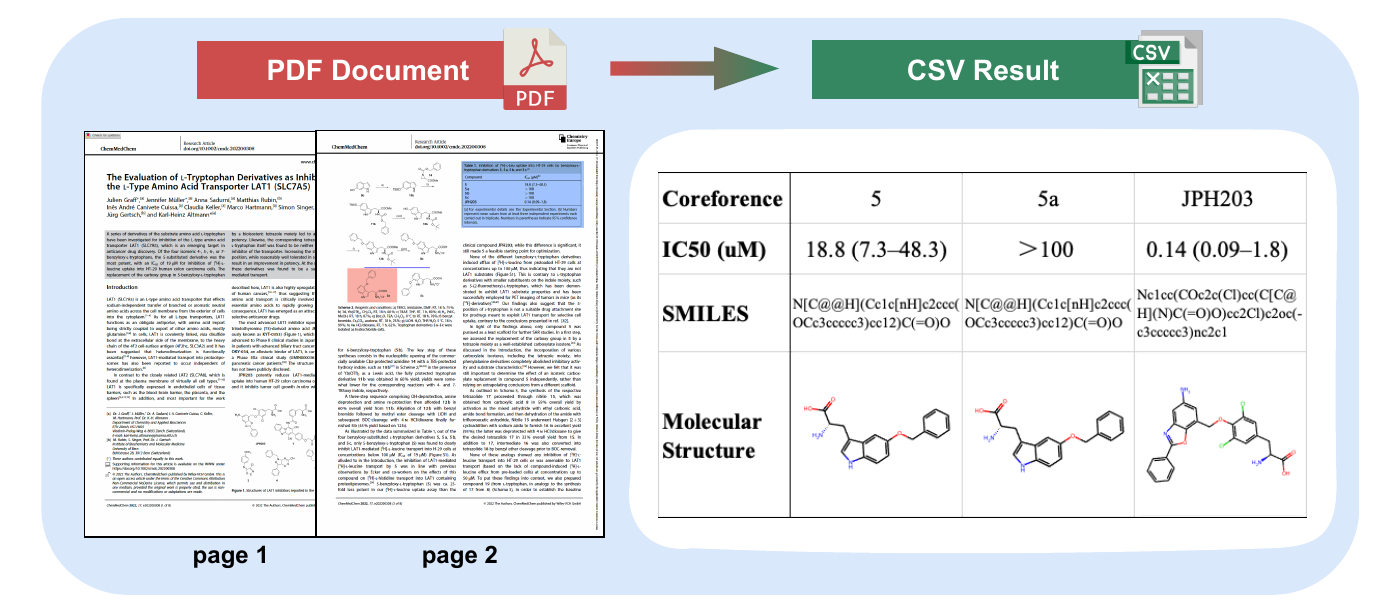}
    \caption{A representative ground truth annotation from the DocSAR-200 benchmark, showing molecular structures linked to activity table entries across multiple pages, formatted as structured CSV records.}
    \label{fig:benchmark_example}
\end{figure}

\section{DocSAR-200 Benchmark}
\label{sec:benchmark}

To systematically evaluate and advance automated extraction of structure–activity relationships (SARs) from heterogeneous scientific sources, we introduce DocSAR-200, a carefully curated benchmark comprising 200 documents (98 patents and 102 research articles) in multiple languages. The design emphasizes three interdependent, hierarchical challenges that any robust extraction system must handle: source diversity, sparse signal localization, and molecular complexity. A concrete example of cross-page molecule-to-table linking is shown in Figure~\ref{fig:benchmark_example}.

\begin{figure*}[t]
    \centering
    \begin{subfigure}[t]{0.32\linewidth}
        \centering
        \includegraphics[width=\linewidth]{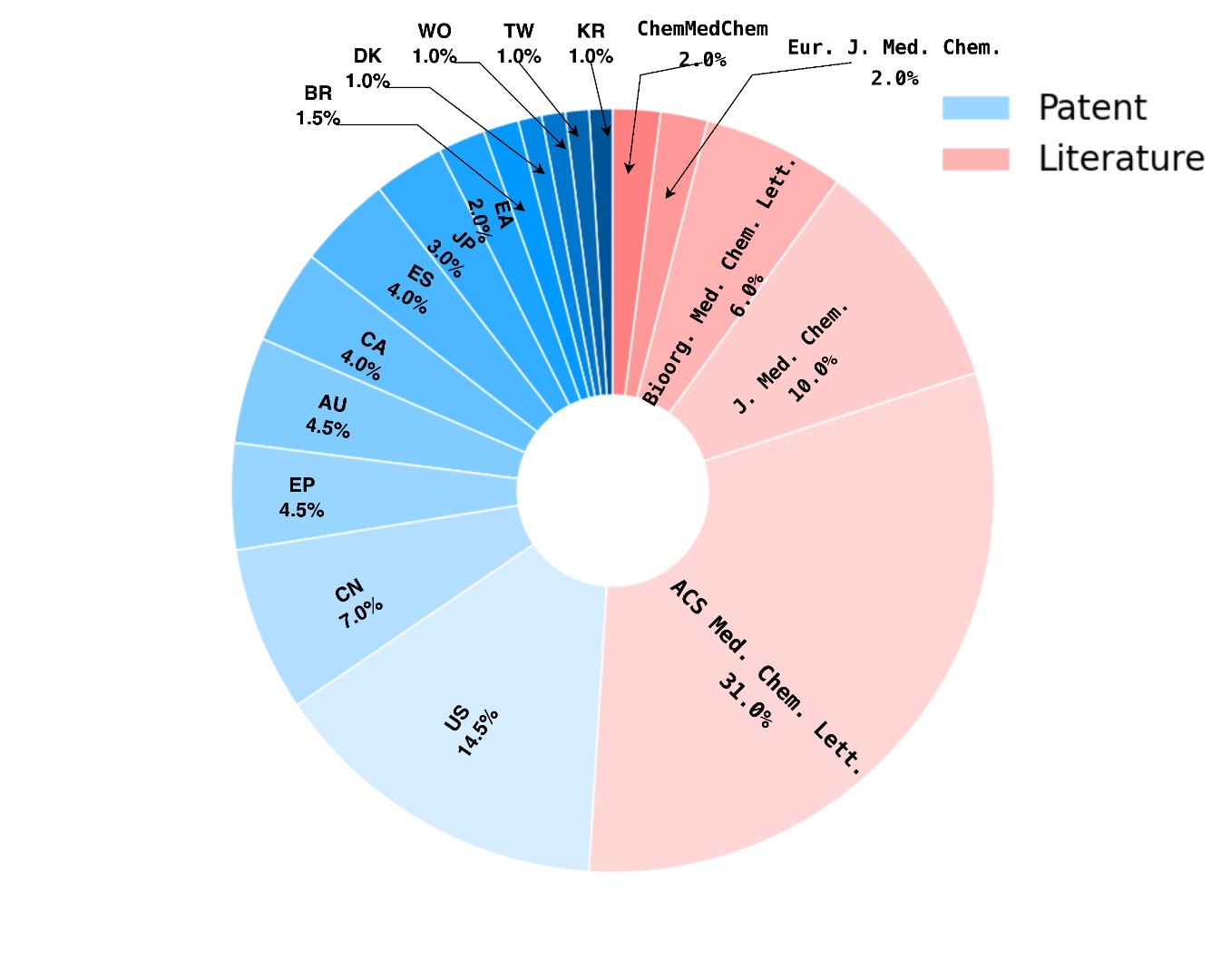}
        \caption{}
        \label{fig:benchmark_overview_a}
    \end{subfigure}
    \hfill
    \begin{subfigure}[t]{0.34\linewidth}
        \centering
        \includegraphics[width=\linewidth]{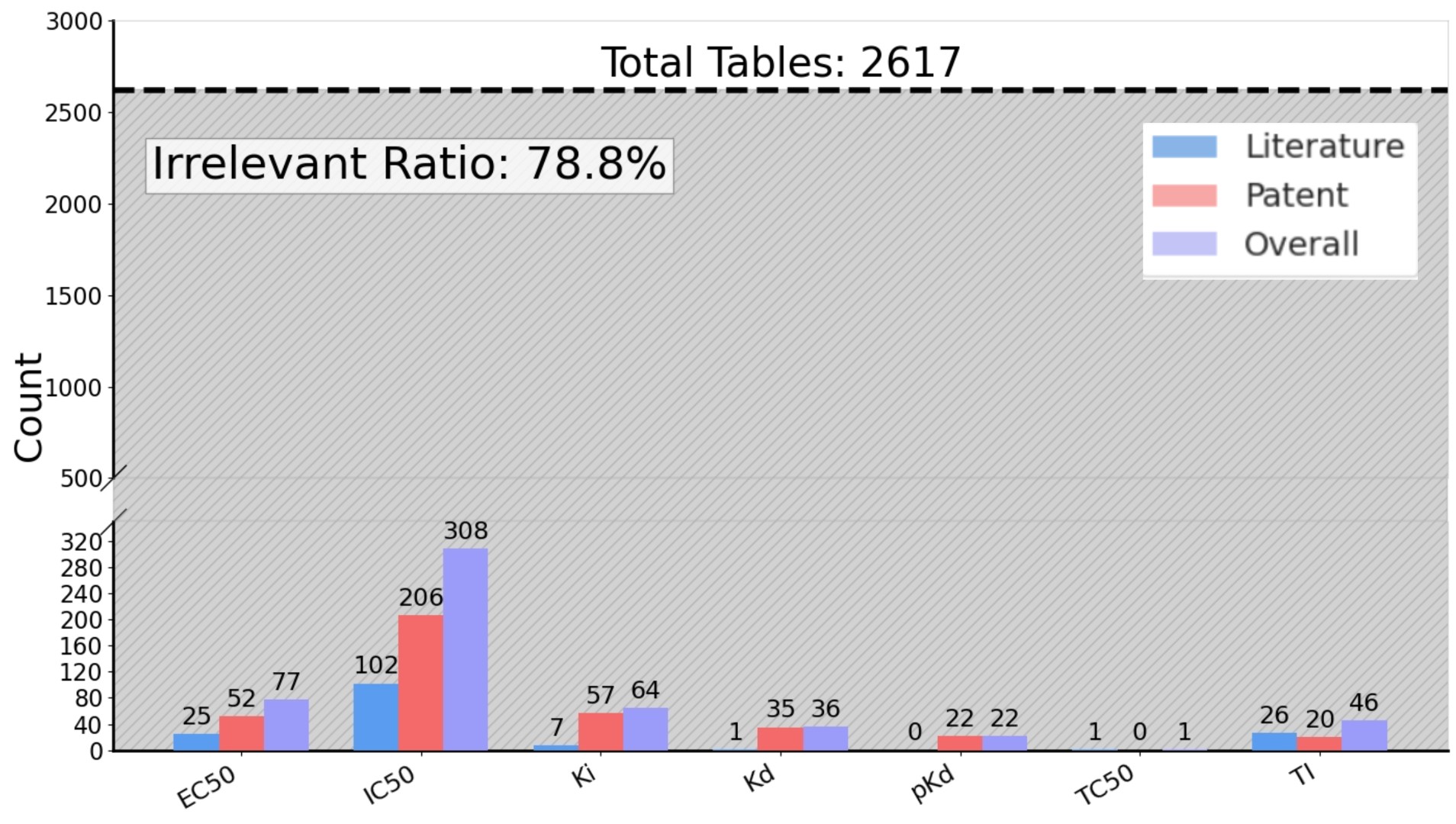}
        \caption{}
        \label{fig:benchmark_overview_b}
    \end{subfigure}
    \hfill
    \begin{subfigure}[t]{0.33\linewidth}
        \centering
        \includegraphics[width=\linewidth]{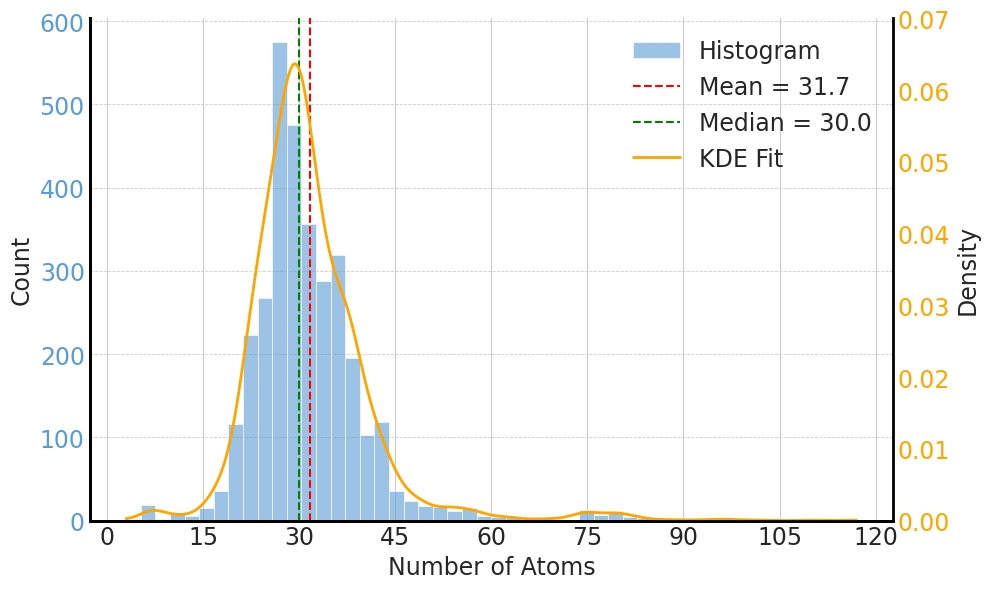}
        \caption{}
        \label{fig:benchmark_overview_c}
    \end{subfigure}
    \caption{Overview of DocSAR-200. (a) File-level composition broken down by patents (blue) and literature (red), with further subdivision across journals and patent offices. (b) Distribution of activity table types within 2617 tables; the gray region indicates the proportion of irrelevant (non-activity) tables. (c) Molecular size distribution measured by atom count, including typical and very large molecules.}
    \label{fig:benchmark_overview}
\end{figure*}

Figure~\ref{fig:benchmark_overview} summarizes the benchmark along three core axes. In \textbf{Figure~\ref{fig:benchmark_overview_a}}, we expose the breadth and heterogeneity of the source corpus: documents span both patents and peer-reviewed literature, originate from multiple patent authorities and journals, and include multilingual content. This multi-type, multi-source coverage reduces overfitting to a single style or domain and reflects real-world variability in formatting and terminology that extraction systems must tolerate; dataset heterogeneity has been shown to be crucial for robust generalization in information extraction benchmarks and downstream models \cite{schwartz2020green, dodge2021available, gao2021scaling}. Literature tends to offer clearer semantic signals and terminology consistency, whereas patents contribute scale and diverse phrasing, creating complementary challenges for parsers.

In \textbf{Figure~\ref{fig:benchmark_overview_b}}, we highlight the difficulty of sparse signal localization. Of the 2617 tables in the benchmark, only a subset encode meaningful activity measurements (e.g., IC$_{50}$, K$_i$, K$_d$), while 77.1\% are irrelevant to activity, yielding a ‘‘needle-in-a-haystack’’ scenario where a system must both distinguish relevant from spurious tabular content and correctly interpret a variety of activity metrics. Extracting sparse, meaningful structured data from noisy or abundant background content is a well-known challenge in scientific and biomedical information extraction, addressed in prior work on distant supervision and table understanding \cite{mintz2009distant, zhong2019pubtabnet, li2022scirex}.

\textbf{Figure~\ref{fig:benchmark_overview_c}} characterizes molecular complexity via the distribution of atom counts across benchmark molecules. The histogram shows a concentration around roughly 30 atoms, with a right-skewed tail extending to substantially larger molecules (including some exceeding 100 atoms). The overlaid kernel density estimate (KDE) provides a smoothed approximation of the empirical distribution, making subtle features (such as skew and potential multi-modality) more visible than a raw histogram alone. The divergence between the mean and median further signals right skew, indicating that a minority of large molecules inflates the average size. KDE is a standard nonparametric technique for continuous distribution estimation that avoids restrictive assumptions and can reveal long tails or structure in molecular property distributions \cite{silverman1986density, hastie2009elements, chen2016comprehensive}.
These size variations require molecular recognition components to be robust across a wide dynamic range, a known consideration in AI-enabled drug discovery where molecular size and complexity influence representation fidelity and predictive performance \cite{vamathevan2019applications, zhou2020artificial}.

Collectively, the three dimensions force extraction pipelines to reconcile heterogeneous source formats, precisely locate sparse but critical activity signals, and handle chemically diverse entities of varying complexity. Detailed annotation methodology and quality control procedures are provided in Appendix~B.

\begin{figure*}
    \centering
    \includegraphics[width=0.94\linewidth]{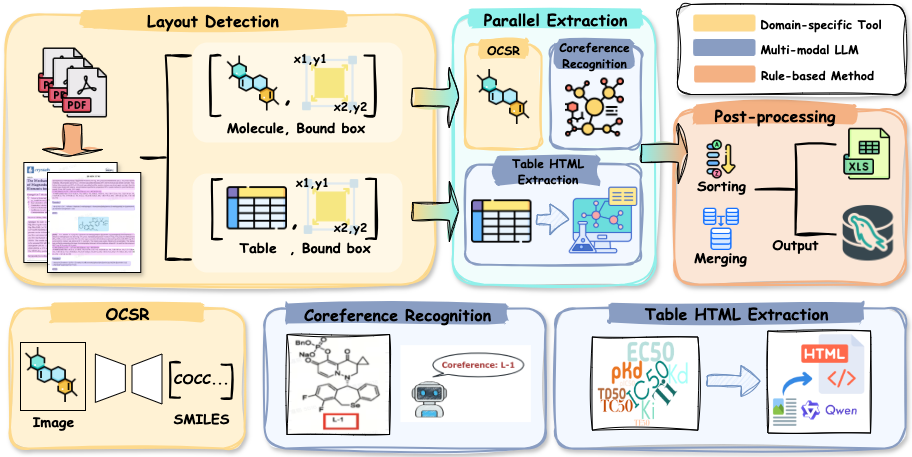}
    \caption{The Doc2SAR framework. The PDF is parsed into layout segments; molecular images and tables are processed in parallel. The framework's strength lies in its synergistic design, using specialized tools like OCSR and fine-tuned MLLMs for perception, and rule-based methods for logical integration.}
    \label{fig:pipeline}
\end{figure*}

\section{Doc2SAR Framework}
\label{sec:framework}

Given an input PDF document \( D \), our goal is to extract a set of relationships \(\mathcal{R} = \{(s_k, a_k)\}_{k=1}^{N}\), where \( s_k \) is a molecule's SMILES string and \( a_k \) is its associated activity. To achieve this, we propose \textbf{Doc2SAR}, a novel synergistic framework that decomposes the complex SAR extraction task into independently optimizable components. Recognizing the inherent brittleness of monolithic end-to-end MLLMs, Doc2SAR strategically integrates specialized domain-specific tools for deterministic tasks (e.g., OCSR) with fine-tuned MLLMs for contextual understanding (e.g., coreference recognition). As illustrated in Figure~\ref{fig:pipeline}, the framework operates through three main stages: Layout Detection, Parallel Extraction, and Post-Processing.

\subsection{Layout Detection}
We employ a YOLO-based object detection model \cite{yolov11} to identify and localize key structural elements within PDF documents. The module converts each PDF page into high-resolution images and extracts spatial coordinates of document elements. For SAR extraction, we focus on two critical categories: molecular structures (\(\mathcal{B}_{\text{mol}}\)) and tabular regions (\(\mathcal{B}_{\text{table}}\)). Each detected element is characterized by bounding box coordinates \((x, y, w, h)\) and confidence scores. The detection function is defined as:
\[
\mathcal{B} = f_{\text{det}}(\mathcal{D}) = \{ \mathcal{B}_{\text{mol}}, \mathcal{B}_{\text{table}} \}
\]

\subsection{Parallel Extraction}
The detected regions are processed through specialized modules that form the core neural perception engine of our framework.

\paragraph{OCSR.} Each molecular image \( I^{\text{mol}}_k \in \mathcal{B}_{\text{mol}} \) is processed by our specialized OCSR model to generate the corresponding SMILES \cite{smiles} string \(s_k\). Our architecture consists of three components: (1) a Swin Transformer \cite{swintrans} image encoder \( f_{\text{enc}} \) for hierarchical visual feature extraction, (2) a two-layer MLP vision-language connector \( f_{\text{conn}} \) following LLaVA \cite{llava} design, and (3) a BART-style \cite{bart} autoregressive decoder \( f_{\text{dec}} \) for SMILES generation. The process is defined as:
\[
s_k = f_{\text{ocsr}}(I^{\text{mol}}_k) = f_{\text{dec}}(f_{\text{conn}}(f_{\text{enc}}(I^{\text{mol}}_k)))
\]
To ensure domain-specific accuracy, we fine-tune the model initialized from MolParser \cite{molparser} on 515 manually curated molecular image-SMILES pairs from patents and literature for 20 epochs with learning rate \(5 \times 10^{-5}\).

\paragraph{Molecular Coreference Recognition.} Establishing correspondence between molecular structure images and their textual identifiers is critical for SAR extraction. For each molecular image \( I^{\text{mol}}_k \), an MLLM analyzes surrounding layout context \( C^{\text{ctx}}_k \) to extract coreference ID \( c^{\text{mol}}_k \). The context encompasses a spatial window (1.5× original dimensions) including captions, nearby text, table content, and figure annotations. The MLLM processes both image and textual context simultaneously, guided by task-specific prompts to identify compound identifiers such as numbers (e.g., "1a", "compound 5") or chemical names. The process handles challenging scenarios including molecules in complex figures and structures with multiple identifiers. This operation is defined as:
\[
c^{\text{mol}}_k = f_{\text{coref}}(I^{\text{mol}}_k, C^{\text{ctx}}_k; \theta_{\text{prompt}})
\]
where \( \theta_{\text{prompt}} \) represents prompt parameters guiding identifier extraction.

\paragraph{Table HTML Extraction.} Scientific documents contain numerous tables, but only a subset contains relevant bioactivity data. We implement a two-stage conditional strategy: first screening for bioactivity keywords (\(\text{IC}_{50}\), \(\text{EC}_{50}\), \(\text{Ki}\), \(\text{Kd}\)), then converting relevant tables to structured HTML. The MLLM handles complex structures including merged cells and multi-level headers while preserving semantic relationships between molecular identifiers and activity values. To ensure quality, the prompt incorporates formatting instructions for numerical precision, cell alignment, and missing data handling. The process is formalized as:
\[
H_j = \begin{cases}
f_{\text{html}}(I^{\text{table}}_j; \theta_{\text{convert}}) & \text{if } f_{\text{screen}}(I^{\text{table}}_j; \theta_{\text{screen}}) = \text{True} \\
\emptyset & \text{otherwise}
\end{cases}
\]

\subsection{Post-processing}
The post-processing module synthesizes parallel extraction outputs through cross-modal alignment using coreference IDs. The alignment operates on molecular data \(\{(s_k, c^{\text{mol}}_k)\}_{k=1}^{M}\) and tabular data \(\{(c^{(j)}_i, a^{(j)}_i)\}_{i=1}^{N_j}\) from each table \(j\). To handle noise and variability in extracted identifiers, we implement multi-tier fuzzy matching based on normalized Levenshtein distance:
\[
\text{sim}(c^{\text{mol}}_k, c^{(j)}_i) = 1 - \frac{\text{Levenshtein}(c^{\text{mol}}_k, c^{(j)}_i)}{\max(|c^{\text{mol}}_k|, |c^{(j)}_i|)}.
\]
The matching employs cascading criteria: exact matching, case-insensitive matching, normalized matching (removing delimiters), and fuzzy matching. A match is valid if similarity exceeds \(\delta = 0.8\), determined empirically. For multiple candidates, we use confidence-based ranking considering similarity scores, spatial proximity, and naming consistency. Quality control includes SMILES validation, activity range checks, and duplicate removal. Successfully linked pairs form structured SAR records:
\[
\mathcal{R} = \{(s_k, a^{(j)}_i) : \max_j \max_i \text{sim}(c^{\text{mol}}_k, c^{(j)}_i) \geq \delta\}.
\]

\subsection{Supervised Fine-Tuning}
While our framework leverages established tools for Layout Detection and OCSR, Table HTML Extraction and Molecular Coreference Recognition require enhanced MLLM capabilities. We employ targeted SFT using a single Qwen2.5-VL-7B-Instruct model \cite{qwen25} on consolidated data combining both sub-tasks. Our consolidated training data comprises:

\begin{itemize}
    \item \textbf{Table Extraction Data}: 10,000 examples randomly sampled from PubTabNet \cite{pubtabnet}, focusing on scientific tables with complex structures similar to those encountered in chemical literature.
    \item \textbf{Molecular Coreference Recognition Data}: 6,373 instances, including 5,552 challenging real-world examples extracted from patents and literature and 821 simpler synthetic cases designed to provide coverage and consistency across less ambiguous layouts. The detailed construction methodology for the synthetic portion is provided in Appendix~C.
    % \textbf{Molecular Coreference Recognition Data}: 6,615 instances including 5,615 real-world examples extracted from patents and literature, plus 1,000 synthetic hard cases covering challenging scenarios with ambiguous identifiers and complex layouts. The detailed construction methodology for synthetic data is provided in Appendix~C.
\end{itemize}

The fine-tuning employs mixed-batch training with examples from both tasks randomly interleaved to promote balanced learning. We train for 3 epochs using batch size 1 and learning rate \(2 \times 10^{-6}\).

\begin{table*}
\centering
\caption{Performance comparison on the DocSAR-200 benchmark, measured by Table Recall (\%).}
\label{tab:main_results}
\begin{tabular}{lcccccc}
\toprule
\multicolumn{1}{c}{\multirow{4}{*}{\textbf{Model}}} & \multicolumn{5}{c}{\textbf{Table Recall (\%)}} \\
\cmidrule(lr){2-6} 
& \multicolumn{1}{c}{\multirow{2.5}{*}{\textbf{Overall}}} & \multicolumn{2}{c}{\textbf{Doc. Type}} & \multicolumn{2}{c}{\textbf{Sub. Task}} \\
\cmidrule(lr){3-4} \cmidrule(lr){5-6}
& & Patents & Literatures & Coref. Rec. & Table. Ex. \\
\midrule
\rowcolor{gray!20}
\multicolumn{6}{c}{\textit{\# End to End Method}} \\
GPT-4o \cite{gpt4o} & 29.30 & 20.75 & 39.19 & 0.00 & 49.23 \\
Gemini-2.5-flash \cite{gemini} & 23.03 & 12.32 & 40.27 & 0.00 & 38.54 \\
Llama3.2-VL-90B-Ins. \cite{llama3} & 4.14 & 0.00 & 23.62 & 0.00 & 7.23 \\
Qwen2.5-VL-72B-Ins. \cite{qwen25} & 28.51 & 16.65 & 45.61 & 0.00 & 47.77 \\
\midrule
\rowcolor{gray!20}
\multicolumn{6}{c}{\textit{\# Doc2SAR Framework}} \\
GPT-4o \cite{gpt4o} & 67.68 & 59.58 & 79.88 & 60.58 & 75.69 \\
Gemini-2.5-flash \cite{gemini} & 70.32 & 64.29 & 78.71 & 58.94 & 78.34 \\
Llama3.2-VL-90B-Ins. \cite{llama3} & 64.32 & 59.57 & 69.92 & 55.37 & 72.18 \\
Qwen2.5-VL-72B-Ins. \cite{qwen25} & 75.55 & 70.67 & 82.52 & 64.07 & 81.03 \\
Qwen2.5-VL-7B-Ins. \cite{qwen25} & 73.76 & 68.27 & 81.95 & 59.03 & 82.62 \\ \rowcolor{gray!10}
\textbf{Ours} & \textbf{80.78} & \textbf{74.09} & \textbf{90.99} & \textbf{65.33} & \textbf{90.08} \\
\bottomrule
\end{tabular}
\end{table*}

\section{Experiments}
\label{sec:experiments}

In this section, we present a series of experiments designed to systematically evaluate our proposed Doc2SAR framework. Our evaluation aims to answer two key research questions: (1) How effective is our synergistic framework approach compared to standard end-to-end methods for SAR extraction? (2) To what extent does SFT contribute to the performance gains?

\subsection{Experimental Setup}
\label{subsec:exp_setup}

\paragraph{Evaluation Metric.}
The primary evaluation is conducted on our DocSAR-200 benchmark, as detailed in Section \ref{sec:benchmark}. We measure performance using Table Recall, a stringent metric that calculates the percentage of data rows where the molecule (identified by its correct SMILES) and all of its associated activity values are extracted and linked perfectly. This metric is chosen for its ability to holistically reflect the practical effectiveness of a method in delivering complete and accurate structured output.

\paragraph{Baseline Models.}
We compare our approach against a suite of powerful, state-of-the-art MLLMs. The baselines include OpenAI's GPT-4o \cite{gpt4o}, Google's Gemini-2.5-flash \cite{gemini}, Meta's Llama3.2-VL-90B-Instruct \cite{llama3}, and Alibaba's Qwen2.5-VL \cite{qwen25} series, specifically the 72B-Instruct and 7B-Instruct versions. All models are evaluated in a zero-shot setting using carefully crafted prompts.

\paragraph{Implementation Details.}

For our experiments, we accessed proprietary models like GPT-4o and Gemini-2.5-flash via their official APIs, while open-source models, including the Llama and Qwen series, were hosted locally from public repositories. To ensure a fair and robust comparison, the task-specific prompts used for all models were developed and optimized on a small, disjoint set of validation documents not present in any of the test sets.

\subsection{Main Results: End-to-End vs. Doc2SAR}
\label{subsec:results_analysis}

This section presents our comprehensive evaluation on the DocSAR-200 benchmark (Table~\ref{tab:main_results}), demonstrating the superiority of our modular framework over end-to-end approaches. Our analysis reveals three key insights: (1) dramatic performance gaps between methodological approaches, (2) framework robustness across diverse document types, and (3) the critical importance of specialized components for complex sub-tasks.

\paragraph{Overall Performance.}
Table~\ref{tab:main_results} reveals a substantial performance gap between our synergistic framework and end-to-end methods. End-to-end approaches show severe limitations: GPT-4o achieves only 29.30\% table recall, while Llama3.2-VL-90B struggles at 4.14\%. In stark contrast, our Doc2SAR framework transforms these same models, with our SFT-enhanced approach achieving 80.78\% table recall, representing a 175\% relative improvement over the strongest end-to-end baseline. Remarkably, our fine-tuned 7B model substantially outperforms the much larger 72B variant (80.78\% vs. 75.55\%), demonstrating that domain-specific training trumps raw model scale for specialized scientific document understanding tasks.

\paragraph{Performance by Document Type.}
Our framework demonstrates robust performance across diverse document formats, with particularly strong results on literature articles (90.99\%) compared to patents (74.09\%). This gap likely reflects structural differences: literature articles feature more standardized table layouts that align well with our PubTabNet training data, while patents exhibit greater format diversity, dense technical language, and complex nested structures that challenge automated processing. The SFT process delivers substantial improvements for literature (81.95\% → 90.99\%), representing an 11\% relative gain that validates our targeted training strategy. The challenging patents highlight the critical value of our synergistic approach, where specialized OCSR and MLLM components collaborate to handle complex, heterogeneous layouts.

% \paragraph{Sub-Task Analysis}
% The sub-task breakdown exposes fundamental limitations of end-to-end approaches. Most critically, \textbf{all end-to-end models achieve 0.00\% on Coreference Recognition}, revealing their inability to perform accurate OCSR followed by cross-modal alignment—a capability essential for SAR extraction. Our modular design addresses this bottleneck through dedicated OCSR tools, achieving 65.33\% on this challenging task. Similarly, for Table Extraction, our approach (90.08\%) dramatically outperforms the best end-to-end method (49.23\%), with an \textbf{83\% relative improvement} that underscores the value of specialized table understanding over generic multimodal processing. These results validate our core hypothesis that complex scientific document understanding requires modular, domain-specific architectures rather than monolithic end-to-end solutions.

\paragraph{Sub-Task Analysis.}
The sub-task breakdown exposes fundamental limitations of end-to-end approaches. Most critically, all end-to-end models achieve 0.00\% on Coreference Recognition, revealing their inability to perform accurate OCSR followed by cross-modal alignment (a complex multi-step process requiring both precise molecular recognition and contextual reasoning capabilities essential for SAR extraction). Our modular design addresses this bottleneck through dedicated OCSR tools, achieving 65.33\% on this challenging task. Similarly, for Table Extraction, our approach (90.08\%) dramatically outperforms the best end-to-end method (49.23\%), with an 83\% relative improvement that underscores the value of specialized table understanding over generic multimodal processing, particularly for scientific tables with domain-specific terminology and complex structures. These results validate our core hypothesis that complex scientific document understanding requires modular, domain-specific architectures rather than monolithic end-to-end solutions.

\subsection{Components Analysis: The impact of SFT}
\label{subsec:key_components_analysis}

To quantify the benefits of our targeted SFT strategy, we conduct isolated evaluations of the two key MLLM-driven components: Molecular Coreference Recognition and Table HTML Extraction. These experiments compare our fine-tuned model against Qwen2.5-VL-7B-Instruct and GPT-4o baseline, demonstrating how domain-specific training enhances capabilities critical to our framework's success. Results are presented in Tables~\ref{tab:coreference_model_comparison} and~\ref{tab:tab_agent_results}.

\paragraph{Coreference Recognition.}
\label{sec:coreference_exp_results}

We evaluate this task on our custom Coreference Recognition Val Set using Recall (\%) as the metric. We categorize test scenarios into \textit{Simple} cases, where molecules appear within structured table environments with clear spatial relationships, and \textit{Hard} cases, where molecules are embedded in unstructured contexts such as figure captions, chemical schemes, or isolated paragraph text. Our SFT-enhanced model achieves an overall recall of 76.40\%, substantially outperforming both GPT-4o (53.40\%) and the base Qwen2.5-VL-7B-Instruct (36.60\%). The improvement is particularly striking for \textit{Hard} scenarios, where our model achieves 70.45\% compared to GPT-4o's 38.81\%, an 82\% relative improvement. Even for \textit{Simple} cases, our model maintains excellent performance at 95.76\%, demonstrating robust capabilities across diverse molecular presentation contexts.

\begin{table}
\centering
\caption{Performance comparison on the Molecular Coreference Recognition task, measured by Recall (\%).}
\label{tab:coreference_model_comparison}
\begin{tabular}{lccc}
\toprule
\multicolumn{1}{c}{\multirow{2}{*}{\textbf{Model}}} & \multicolumn{3}{c}{\textbf{Recall (\%)}}           \\  
\cmidrule{2-4}
\multicolumn{1}{c}{}                                & \textbf{Overall} & \textbf{\textit{Simple}} & \textbf{\textit{Hard}}    \\  
\midrule
GPT-4o                 & 53.40 & \textbf{99.39} & 38.81 \\
Qwen2.5VL$^{\dagger}$     & 36.60 & 88.48          & 17.91 \\ 
\rowcolor{gray!10}
\textbf{Ours}          & \textbf{76.40} & 95.76          & \textbf{70.45} \\
\bottomrule
\end{tabular}
\\[0.5em]
\raggedright{\small $^{\dagger}$Qwen2.5-VL-7B-Instruct}
\end{table}

\paragraph{Table HTML Extraction.}
\label{subsec:table_extraction_exp}

We evaluate this task on the standard PubTabNet Val Set using the TEDS score (\%) metric. Tables are categorized as \textit{Simple} (regular grid layouts with uniform cell structures and clear row-column organization) and \textit{Hard} (complex layouts featuring merged cells, multi-level headers, nested structures, or irregular formatting). Our model achieves a superior overall TEDS score of 58.29\%, outperforming GPT-4o (48.05\%) and the base model (43.51\%). The performance gain is most pronounced for \textit{Hard} tables, where our model scores 54.44\% versus GPT-4o's 43.31\%, a 26\% relative improvement. This enhanced structural understanding of complex table layouts is crucial for accurately processing the diverse tabular formats found in scientific literature.

\section{Case Study}

To provide qualitative evidence of our framework's real-world efficacy, we present two case studies showcasing intra-page vs. cross-page association and monolingual vs. multilingual document processing. These cases demonstrate Doc2SAR's ability to handle diverse layouts and correctly associate molecular structures with activity data.

\begin{table}
\centering
\caption{Performance on the Table HTML Extraction task, measured by average TEDS score (\%).}
\label{tab:tab_agent_results}
\begin{tabular}{lccc}
\toprule
\multicolumn{1}{c}{\multirow{2}{*}{\textbf{Model}}} & \multicolumn{3}{c}{\textbf{Avg. TEDS score (\%)}}           \\  \cmidrule{2-4}
\multicolumn{1}{c}{}                                & \textbf{Overall} & \textbf{\textit{Simple}} & \textbf{\textit{Hard}}    \\  \midrule
GPT-4o                                              & 48.05            & 52.57                   & 43.31                   \\
Qwen2.5VL$^{\dagger}$                                  & 43.51            & 49.24                   & 37.21                   \\  \rowcolor{gray!10}
\textbf{Ours}                                       & \textbf{58.29}   & \textbf{61.72}          & \textbf{54.44}   \\
\bottomrule
\end{tabular}
\\[0.5em]
\raggedright{\small $^{\dagger}$Qwen2.5-VL-7B-Instruct}
\end{table}

\begin{figure}
\centering
\includegraphics[width=\linewidth]{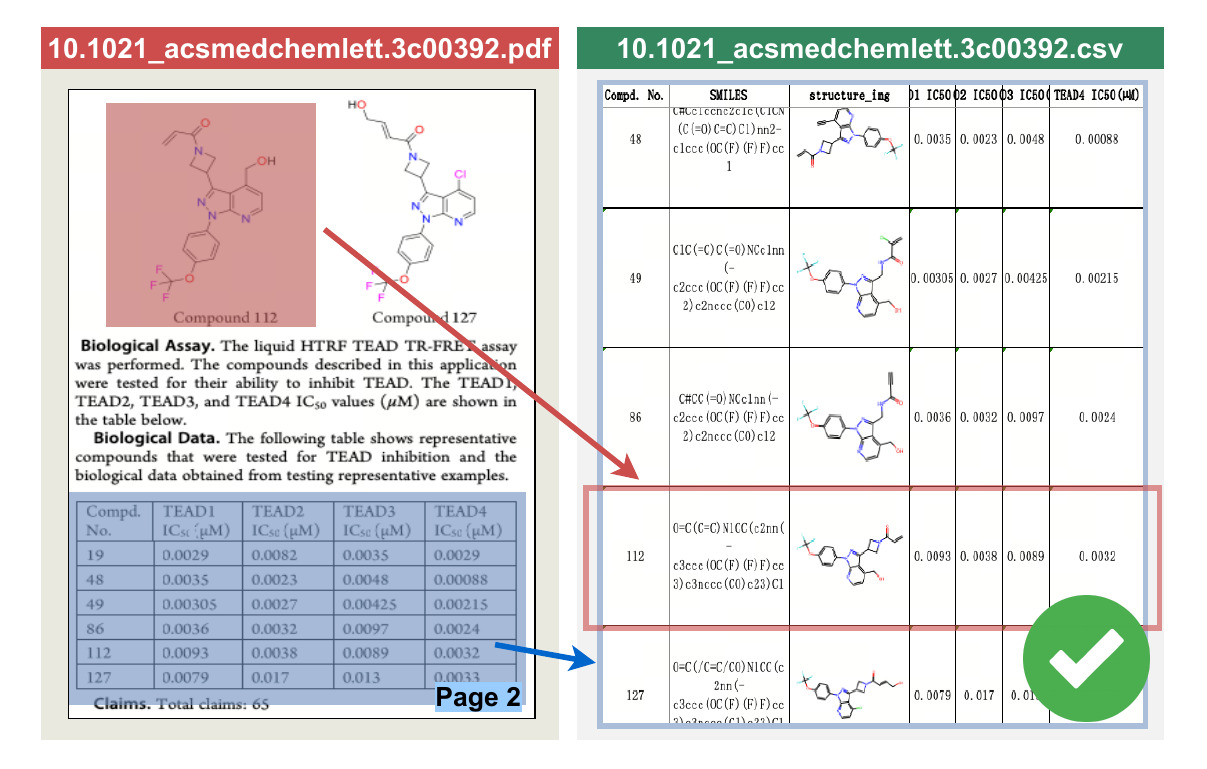}
\caption{Case study in ACS Medicinal Chemistry Letters \texttt{10.1021\_acsmedchemlett.3c00392}. Note that the red and blue boxes (masks) in the case are only for auxiliary visualization display and are not included in the document. Doc2SAR correctly performs intra-page association, linking a molecular structure to its activity table on the same page.}
\label{fig:case1}
\end{figure}

The first case (Figure~\ref{fig:case1}) involves a biomedical research paper where the molecular structure and activity data are co-located on the same page. Our framework successfully performs this intra-page association by: (1) extracting the molecular structure via OCSR, (2) identifying the compound's textual identifier through coreference recognition, and (3) locating the matching table row. The system correctly extracts the IC$_{50}$ value (2.3 nM), while GPT-4o failed due to OCSR errors.

\begin{figure}
\centering
\includegraphics[width=\linewidth]{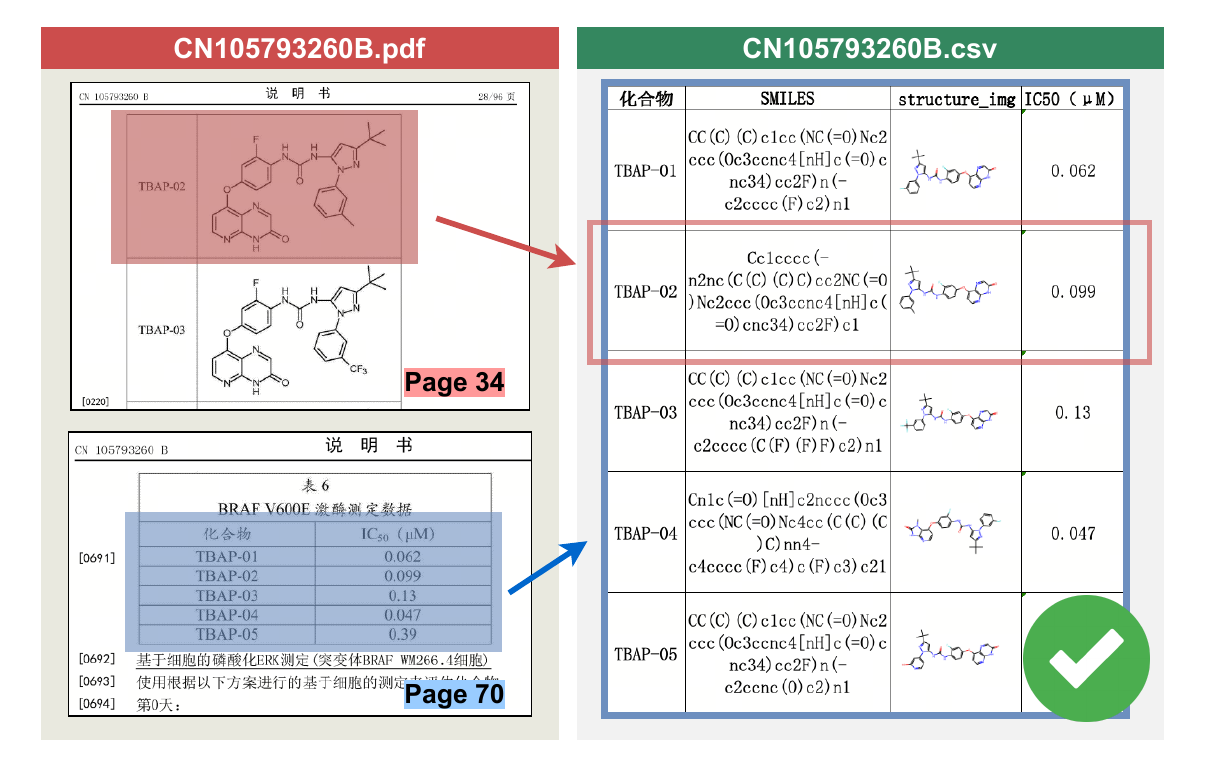}
\caption{Case study in C.N. patent \texttt{CN105793260B}. Note that the red and blue boxes (masks) in the case are only for auxiliary visualization display and are not included in the document. Doc2SAR demonstrates its robustness by performing a long-range, cross-page association, linking a molecule on Page 34 with its activity data on Page 70.}
\label{fig:case2}
\end{figure}

The second case (Figure~\ref{fig:case2}) presents a challenging scenario from a Chinese patent, where the molecular structure appears on Page 34 while activity data resides in a table 36 pages later. This cross-page span, combined with patent formatting and non-English text, overwhelms end-to-end approaches. Our framework demonstrates robust performance by accurately recognizing the molecular structure, extracting the compound identifier via multilingual coreference recognition, and successfully linking across the 36-page gap using fuzzy matching (similarity score: 0.95). This demonstrates scalability and language-agnostic capabilities.

For a more comprehensive analysis, including additional successful extractions and a discussion of challenging failure cases, please refer to Appendix~D.

% The first case, illustrated in Figure~\ref{fig:case1}, involves a typical scientific literature article where the molecular structure and its corresponding activity data are located on the same page. Our framework successfully performs this intra-page association, correctly identifying the molecule, locating its entry in the activity table, and extracting the complete data row.

% The second case, shown in Figure~\ref{fig:case2}, highlights a more challenging scenario from a Chinese patent. The molecular structure is found on Page 34, while its activity data is in a table 36 pages later on Page 70. This large, cross-page span poses a significant challenge for monolithic end-to-end models. In contrast, our modular Doc2SAR framework excels, robustly linking the disparate information via its coreference-based merging strategy. This case also showcases the framework's versatility in handling non-English documents.

\section{Related Work}

\paragraph{Chemical Information Extraction.}
Early rule‐based systems could recognise chemical names or parse fixed reaction templates, but generalised poorly to diverse layouts.  
Recent efforts have shifted toward neural, modality‐specific extractors.  
For patents, the \textsc{ChEMU}2022 shared task released a benchmark for chemical NER and event extraction with layout cues~\cite{Li2022ChEMU}.  
Transformer OCSR models include DECIMER.ai with image–to–sequence decoding~\cite{Rajan2024DECIMER}, MolScribe's image-to-graph generation~\cite{Qian2023}, and MolGrapher's detection-plus-graph pipeline~\cite{Morin2023MolGrapher}. These models now achieve more than 90\% top-1 exact match on curated sets.  
PatCID further demonstrated large-scale extraction of 80 M structure images from patents by combining segmentation with OCSR~\cite{Morin2024PatCID}.  
However, these systems operate on a single input type (text \emph{or} image) and therefore cannot recover cross-modal SARs.

\paragraph{Domain Large Language Models.}
Large language models are increasingly adapted for chemistry and materials science.  
ChemCrow~\cite{Bran2023} couples GPT-4 with cheminformatics tools to plan multi-step syntheses;  
Vangala~\textit{et al.}~\cite{Vangala2024} fine-tuned GPT-J to mine patent reactions, boosting recall by 26 \% over grammar-based baselines;  
FuncFetch~\cite{Smith2024} uses GPT-3.5 to extract $>$ 20 k enzyme–substrate pairs with high precision;  
and SynAsk~\cite{Zhang2025} combines a knowledge base with a GPT-derived model for interactive retrosynthesis and reaction-condition suggestions.  
Although these LLMs excel at text-centric tasks, evaluations show degraded accuracy once molecular diagrams or numeric tables must be aligned with textual mentions~\cite{Dagdelen2024}.  
Doc2SAR addresses this limitation by integrating fine-tuned multimodal LLMs with domain-specific tools to link text, figures and tables in documents.

\paragraph{Multimodal Documents Understanding.}
Vision language document models fuse layout and visual cues: LayoutLMv3~\cite{Huang2022} unifies text–image masking for pre-training,  
UReader~\cite{Ye2023} adopts an OCR-free encoder for long documents,  
and DocOwl 1.5~\cite{Hu2024} refines multimodal alignment for page-level QA.  
Benchmarks such as UniMMQA~\cite{Luo2023MMQA} and SPIQA~\cite{Pramanick2024} probe reasoning across text, tables and figures, yet contain few chemical diagrams and no ground-truth SAR links.  
Therefore, DocSAR-200 contributes the first dataset in which molecular structures and activity tables are jointly annotated, and Doc2SAR provides a framework that demonstrably outperforms general VLMs on this document-level SARs extraction.

% \section{Conclusion}
% \label{sec:conclusion}

% We present \textbf{DocSAR-200}, a new benchmark of 200 diverse scientific documents designed to advance automated SAR extraction. Alongside, we introduce \textbf{Doc2SAR}, a novel \textbf{synergistic framework} that integrates domain-specific tools with SFT-enhanced MLLMs. Our approach achieves a state-of-the-art Table Recall of \textbf{80.78\%}, outperforming the strongest end-to-end GPT-4o baseline by \textbf{51.48\%}, underscoring the strength of modular designs and the critical role of task-specific fine-tuning.

% While Doc2SAR marks a significant step forward, several avenues remain for future exploration:

% \begin{itemize}
%     \item \textbf{Benchmark Expansion:} Extend DocSAR-200 to include more languages (e.g., Chinese, German), scientific fields, and complex annotation types.
%     \item \textbf{Methodological Advances:} Explore hybrid models and more data-efficient fine-tuning to reduce reliance on curated SFT data.
%     \item \textbf{Broader Applications:} Adapt the framework to other scientific extraction tasks such as experimental protocols and material properties.
% \end{itemize}

% We hope DocSAR-200 serves as a foundation for continued research in scientific document understanding, and that Doc2SAR provides a strong, adaptable baseline for the community to build upon.

\section{Conclusion}
\label{sec:conclusion}

In this work, we introduce DocSAR-200, a benchmark of 200 scientific documents, and Doc2SAR, a synergistic framework integrating domain-specific tools with SFT-enhanced MLLMs for automated SAR extraction. Doc2SAR achieves state-of-the-art Table Recall of 80.78\%, a 175\% relative improvement over the strongest end-to-end baseline, demonstrating that specialized architectures outperform monolithic approaches.

Our synergistic framework demonstrates that strategic integration of specialized components yields superior performance over large general-purpose models. By enabling each component to excel at its designated task while maintaining seamless cross-modal coordination, this modular paradigm fundamentally addresses the limitations inherent in monolithic approaches to scientific document understanding.

Future work will explore extending multilingual capabilities, developing data-efficient methods, and adapting the framework to other tasks. We anticipate DocSAR-200 will catalyze research in scientific document understanding.

\bibliography{aaai25}
% \input{checklist}

% Appendix
\newpage
\appendix
\section{Prompts Design}
\label{appendix:prompts}

This section details the specific prompts within our Doc2SAR framework. These prompts were carefully designed to guide the models in performing their specialized sub-tasks: extracting activity data from table images into HTML format, and recognizing molecular coreferences from molecular structure images, distinguishing between cases where the structure is found outside or inside a table. 

\begin{prompt}{Prompt for extracting activity data from table image}
\textbf{\small System Message:}
You are an expert in the field of pharmaceutical chemistry, and your task is to extract activity data from an image in tabular HTML format.
\\
\\
\textbf{\small User Message:}\\
First, analyze the table content from the provided image and convert it into an HTML format. The HTML should include all rows and columns, representing the table structure exactly as it appears in the image.\\
Use the following guidelines:\\
1. Determine if the table has a header row. If the first row is a header, represent its cells with \textasciigrave \textless th\textgreater \textasciigrave\ elements. Otherwise, represent all rows (including the first row) with \textasciigrave \textless td\textgreater \textasciigrave\ elements.\\
2. Include a \textless table\textgreater\ element containing all rows (\textasciigrave \textless tr\textgreater \textasciigrave) and cells (\textasciigrave \textless th\textgreater \textasciigrave\ for headers, \textasciigrave \textless td\textgreater \textasciigrave\ for data).\\
3. Retain the exact content of each cell, including special characters, numerical values, text and even empty cell.\\
4. If a cell contains molecular structure images, replace it with the token "[mol]".\\
5. If the table contains merged cells, use appropriate attributes like \textasciigrave rowspan \textasciigrave\ or \textasciigrave colspan \textasciigrave\ to reflect the merging.\\
6. Pay attention to the following content: \\
(1) Attribute: ['EC50', 'IC50', 'Ki', 'Kd', 'pKd', 'TD50', 'Ti', 'TC50']\\
(2) Unit: ['uM', 'nM', '\%', 'kcal/mol']\\
If none of these Attribute keywords are found, it is not an activity table. Your entire output should be the exact string: \textless table\textgreater None\textless/table\textgreater
\\
\\
\textbf{\small Example HTML structure:}\\
\textasciigrave \textasciigrave \textasciigrave html\\
\textless table\textgreater\\
\textless tr\textgreater\\
\textless th\textgreater xxx\textless/th\textgreater\\
\textless th\textgreater xxx\textless/th\textgreater\\
\textless th\textgreater xxx\textless/th\textgreater\\
\textless/tr\textgreater\\
\textless tr\textgreater\\
\textless td\textgreater xxx\textless/td\textgreater\\
\textless td\textgreater x\textless/td\textgreater\\
\textless td\textgreater x\textless/td\textgreater\\
\textless/tr\textgreater\\
\textless tr\textgreater\\
\textless td\textgreater xxx\textless/td\textgreater\\
\textless td\textgreater x\textless/td\textgreater\\
\textless td\textgreater xx\textless/td\textgreater\\
\textless/tr\textgreater\\
...\\
\textless/table\textgreater\\
\textasciigrave \textasciigrave \textasciigrave
\\
\\
\textbf{\small [Image of a table may contain activity data]}
\end{prompt}

\begin{prompt}{Prompt for recognizing coreference from molecular structure image (In Case of Outside the table)}
\textbf{\small System Message:}
You are an expert in the field of pharmaceutical chemistry, and your task is to recognize the coreference from a molecular structure image in JSON format.
\\
\\
\textbf{\small User Message:}\\
I have a molecular structure image **not inside a table**.\\
Your task is to **extract only the molecular coreference** associated with the molecule. The coreference is typically a unique identifier (e.g., numeric, alphabetic, or alphanumeric code) found in the image.\\
Important Instructions:\\
1. **Focus on coreference priority**:\\
 - **First, check the left side of the molecule** for a valid coreference.\\
 - **If the left side is empty or invalid, check the right side.**\\
 - Do **not** check inside any table.\\
2. **Ignore the molecular structure itself**:\\
 - Do not output SMILES, InChI, or any chemical structure representation.\\
 - Focus only on extracting the **molecular coreference**.\\
3. Strictly Follow The JSON Output Format
\\
\\
\textbf{\small Example JSON structure:}\\
\textasciigrave \textasciigrave \textasciigrave json\\
\{\{"compound\_id": "\textless Molecular coreference or [None]\textgreater"\}\}\\
\textasciigrave \textasciigrave \textasciigrave
\\
\\
\textbf{\small [Image of a molecular structure containing corresponding coreference]}
\label{prompt_out_table}
\end{prompt}

\begin{prompt}{Prompt for recognizing coreference from molecular structure image (In Case of Inside the table)}
\textbf{\small System Message:}
You are an expert in the field of pharmaceutical chemistry, and your task is to recognize the coreference from a molecular structure image in JSON format.
\\
\\
\textbf{\small User Message:}\\
I have a molecular structure image located **inside a table**.\\
Your task is to **identify only the molecular coreference** in the image, which is typically a unique identifier (e.g., a numeric, alphabetic, or abbreviation) associated with the molecule in the image.\\
Important Instructions:\\
1. **Priority of Coreference Extraction**:\\
 - The molecular coreference **is usually inside the same table cell** as the molecular image.\\
 - If the table cell is empty or ambiguous, check nearby table cells.\\
 - Do **not** look outside the table.\\
2. **Ignore the molecular structure itself**:\\
 - Do not output SMILES, InChI, or any chemical structure representation.\\
 - Focus only on extracting the **molecular coreference**.\\
3. Strictly Follow The JSON Output Format
\\
\\
\textbf{\small Example JSON structure:}\\
\textasciigrave \textasciigrave \textasciigrave json\\
\{\{"compound\_id": "\textless Molecular Coreference or [None]\textgreater"\}\}\\
\textasciigrave \textasciigrave \textasciigrave
\\
\\
\textbf{\small [Image of a molecular structure containing corresponding coreference]}
\label{prompt_in_table}
\end{prompt}

\begin{figure*}[!ht]
    \centering
    \includegraphics[width=\linewidth]{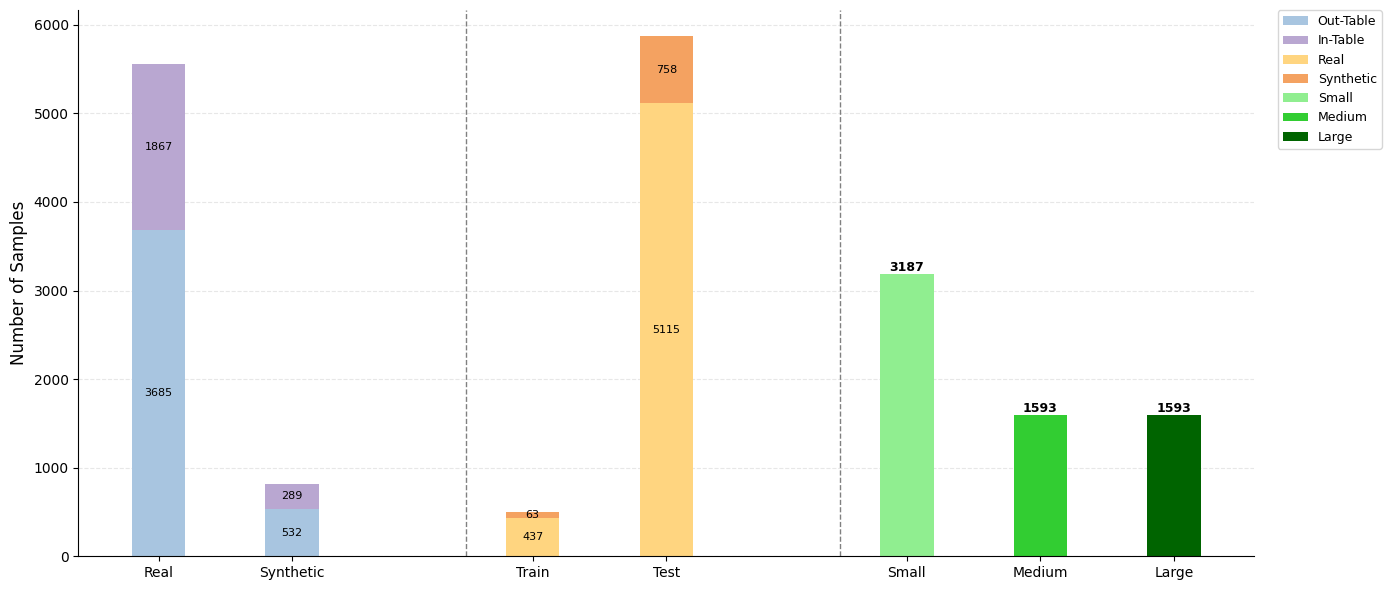} % Adjust width as needed
    \caption{Composition of the SFT dataset. Left: breakdown of real and synthetic samples by layout type (in-table vs. out-of-table). Center: train (500 samples) and test (5,873 samples) splits, showing contributions from real and synthetic sources. Right: image size distribution categorized into Small, Medium, and Large, with respective sample counts.}
    \label{fig:sft_data_stats}
\end{figure*}

\section{Annotation Details and Quality Control}
\label{appendix:annotation}

The DocSAR-200 annotation process was designed to ensure high fidelity and reproducibility. Each document was processed through a two-stage pipeline: first, GPT-4o’s multimodal capabilities
% ~\cite{gpt4o}
were employed to extract preliminary tabular information (e.g., compound names, activity values) into structured formats. In parallel, molecular structure images were located using coreference links or positional heuristics.

Subsequently, trained annotators reviewed and corrected all entries, ensuring accurate alignment between molecules, labels, and extracted tables. Molecular images were parsed into SMILES representations via open-source OCSR tools, such as \texttt{MolParse}
% ~\cite{molparser}
, followed by human validation.

\textbf{Challenge considerations.} Several challenges were addressed during annotation. Multi-page layouts—common in patents—were handled using persistent compound IDs and bounding box coordination to track dispersed molecular and activity references. Markush structures, which cannot be resolved into single molecules, were excluded. For multilingual documents (English, Chinese, Japanese), language-specific tokenization and layout normalization were applied. Native speakers ensured semantic consistency across languages.

\textbf{Validation protocols.} All SMILES were validated using \texttt{RDKit}
% ~\cite{RDKitWebsite}
to ensure chemical correctness and canonicalization. Heuristic filters removed malformed structures and disconnected fragments. Coordinates and page indices were manually verified to prevent drift or misalignment.

\textbf{Annotation statistics.} The finalized dataset contains \textbf{3329 valid SMILES} across \textbf{200 documents} (98 patents, 102 research articles) in three languages. The annotation process required over \textbf{160 hours} of expert effort.

These measures collectively ensure that DocSAR-200 offers a reliable, high-quality benchmark that reflects the real-world complexity of SAR extraction.

\section{SFT Dataset Construction Details}
\label{appendix:sft_details}

The SFT dataset is meticulously constructed from a combination of real-world and synthetic molecular images, designed to enhance the model's capabilities in recognizing molecular coreferences. To create a unified training corpus, all samples are processed into one of two standardized layouts: an \textbf{in-table layout}, which captures the entire table row containing the target molecule, and an \textbf{out-of-table layout}, where the molecule image is cropped with fixed padding to preserve essential surrounding context.

The real data samples originate from a diverse collection of patents and scientific PDFs, reflecting authentic, ``in-the-wild'' scenarios. Our semi-automated annotation pipeline begins with \texttt{UniFinder} to detect the bounding boxes of molecules, followed by \texttt{PyMuPDF} 
% \cite{PyMuPDF_docs}
to crop the corresponding images based on their layout type. An initial annotation pass is performed by \texttt{Qwen2.5-VL-7b-Instruct} 
% \cite{qwen25}
using domain-specific prompts (see Appendix~\ref{prompt_in_table}), with all generated labels subsequently undergoing a rigorous two-person verification process to ensure accuracy. To further augment the dataset, especially for covering edge cases and creating challenging negative samples, we generate synthetic data. These samples are created by rendering molecular structures from PubChem 
% \cite{Kim2025PubChemUpdate}
into both layouts. Their labels are randomly generated based on real-data formats, and for out-of-table samples, label positions are programmatically assigned using OpenCV 
% \cite{opencv_library}
to approximate empirical distributions (e.g., bottom: 50\%, top: 30\%).

The final consolidated SFT dataset comprises a total of \textbf{6,373} samples. This includes \textbf{5,552} real-world samples (3,685 in in-table layout and 1,867 out-of-table) and \textbf{821} synthetic samples (532 in-table and 289 out-of-table). The dataset is further divided into \textbf{437} training samples and \textbf{5,873} testing samples. 

Additionally, we analyze image size distributions to better understand the dataset characteristics. As shown in Figure~\ref{fig:sft_data_stats}, image sizes are categorized into three groups: \textbf{Small} (0–598,590 px\textsuperscript{2}, 3,187 samples), \textbf{Medium} (598,590–927,368 px\textsuperscript{2}, 1,593 samples), and \textbf{Large} (>927,368 px\textsuperscript{2}, 1,593 samples). The image sizes range from a minimum of 191,205 px\textsuperscript{2} to a maximum of 2,098,539 px\textsuperscript{2}, with a mean of 668,888.55 px\textsuperscript{2}. Figure~\ref{fig:sft_data_stats} provides a comprehensive visualization of the dataset statistics, including real/synthetic distributions, train/test splits, and image size categories.

\begin{figure*}[!htb]
\centering
\includegraphics[width=\linewidth]{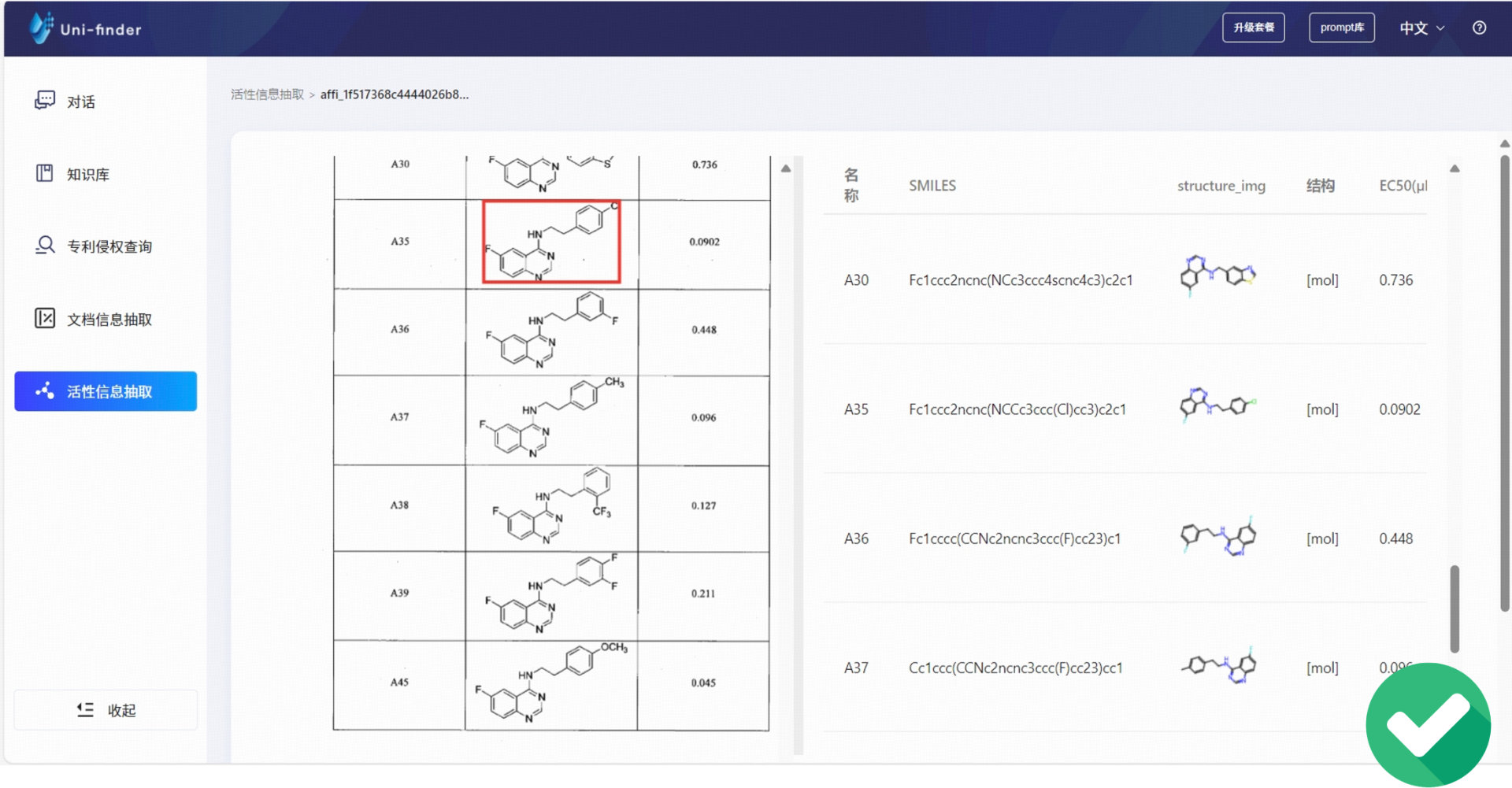}
\caption{A successful extraction case from patent \texttt{CN102574816A}. Doc2SAR accurately processes the provided table, correctly extracting the molecular structures, their coreference IDs (e.g., A35, A36), and associated activity values into a clean, structured output.}
\label{fig:case3}
\end{figure*}

\begin{figure*}[!htb]
\centering
\includegraphics[width=\linewidth]{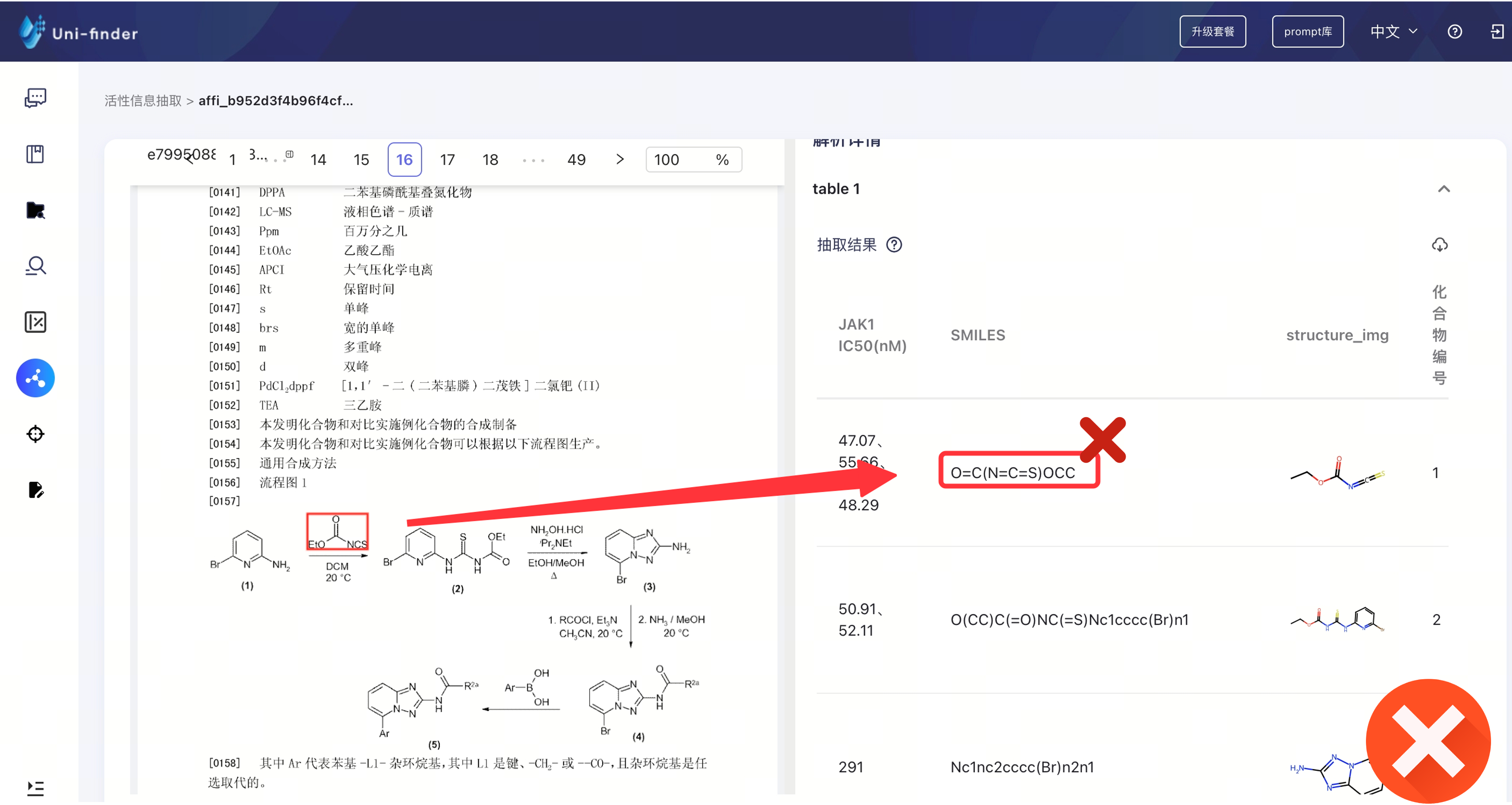}
\caption{A challenging failure case from literature \texttt{CN102482273B}. The system incorrectly extracts the SMILES string for the molecule with coreference '1'. This misidentification may be attributed to the complexity of the reaction scheme diagram, where the simple label '(1)' refers to the initial reactant but is located within a visually complex multi-step synthesis context, leading to an incorrect association.}
\label{fig:case4}
\end{figure*}

\paragraph{Data Quality Control.} All real-data annotations underwent double labeling and third-party adjudication, supported by automated checks on field integrity, SMILES validity, and spatial alignment. For synthetic data, layout logic and label formats were systematically validated to ensure consistency with observed real-world patterns. This combination of rigorous human review and controlled synthesis maintained both precision and diversity within the SFT corpus.

\section{Case Study}
\label{appendix:case_studies}

% This section presents additional case studies to provide further insight into our system's performance on diverse, real-world documents. We showcase a representative successful extraction from a patent and a challenging case involving a recognition error to offer a transparent view of Doc2SAR's current capabilities and areas for future improvement.

To further illustrate the capabilities and limitations of Doc2SAR, this appendix presents additional case studies. We showcase two representative examples below. The first is a successful extraction from a patent (Figure~\ref{fig:case3}), demonstrating the framework's robustness in accurately processing well-structured tables with embedded molecular images. In contrast, the second case presents a failure from a scientific article (Figure~\ref{fig:case4}) that highlights a key challenge: the misidentification of molecules within complex reaction schemes. This transparent analysis provides insight into both the system's current strengths and its areas for future development.

\newpage
\section{Web Application}
\label{appendix:webapp}

To demonstrate the practical utility and facilitate user interaction with our Doc2SAR system, we have developed an intuitive web application. This platform serves not only as a showcase for the end-to-end extraction capabilities of Doc2SAR but also as a tool for users to verify results and understand the system's performance on their own documents. The application allows users to easily upload PDF documents (scientific articles or patents) and submit them for automated molecular activity information extraction.

\begin{figure*}[ht]
    \centering
    % First Row
    \begin{subfigure}{0.48\textwidth}
        \centering
        \includegraphics[width=0.98\linewidth]{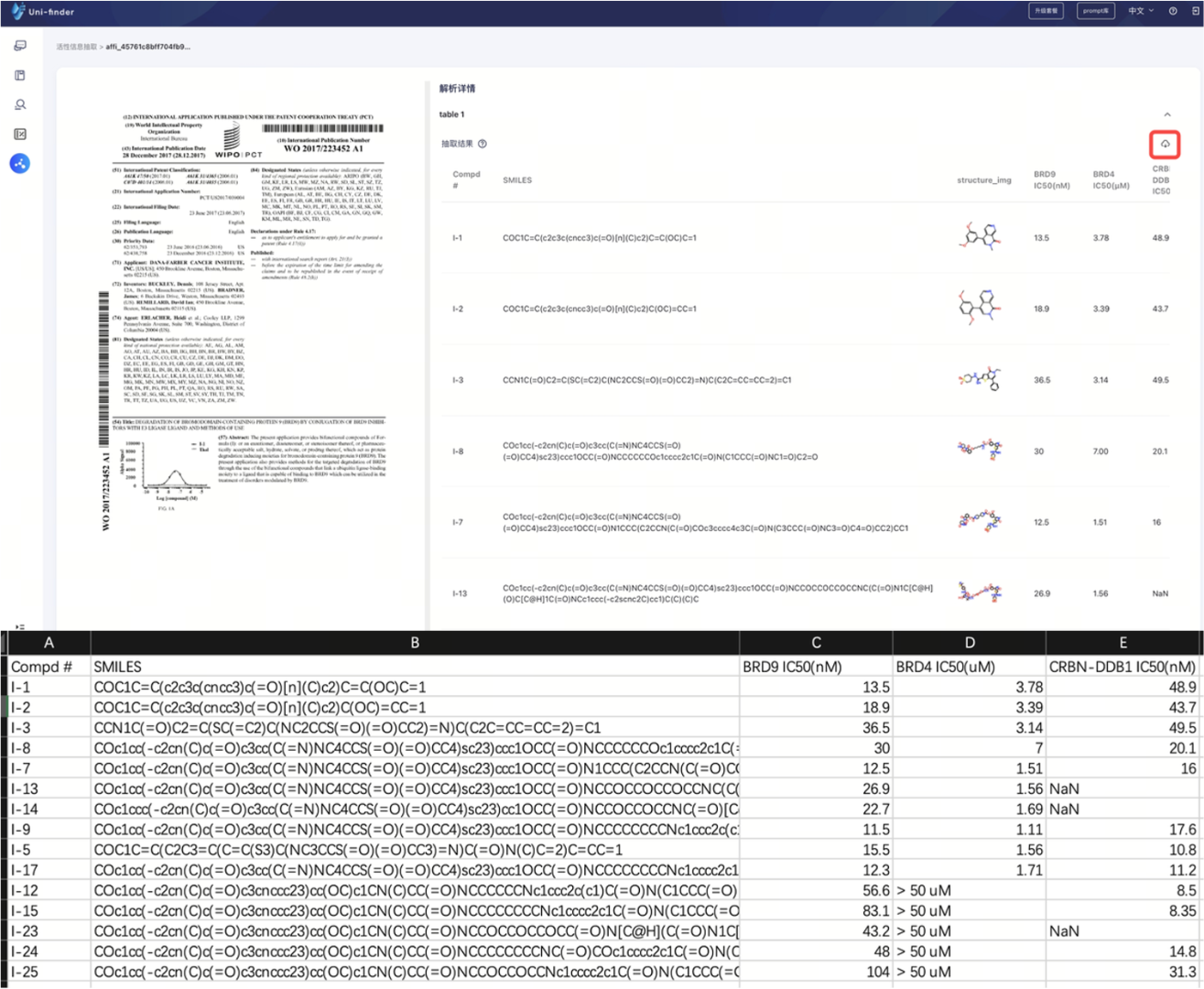}
        \caption{}
    \end{subfigure}\hfill
    \begin{subfigure}{0.48\textwidth}
        \centering
        \includegraphics[width=0.98\linewidth]{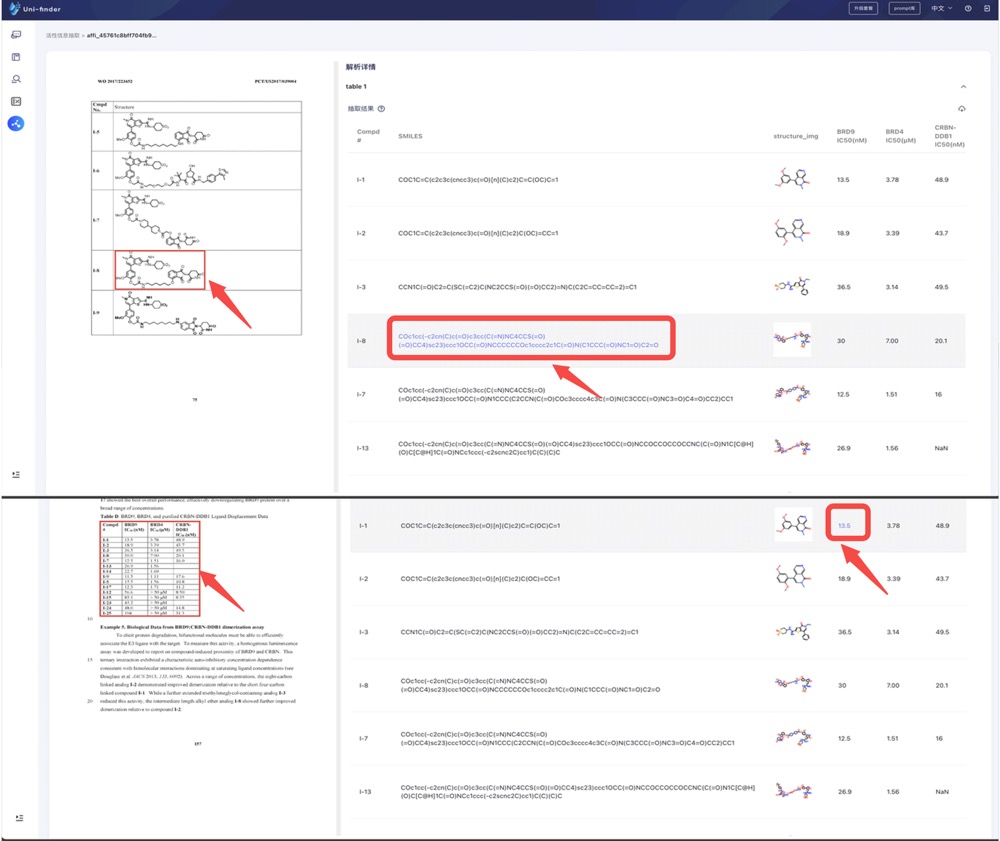}
        \caption{}
    \end{subfigure}

    \vspace{1ex} % Adds a little vertical space between rows

    % Second Row
    \begin{subfigure}{0.48\textwidth}
        \centering
        \includegraphics[width=0.98\linewidth]{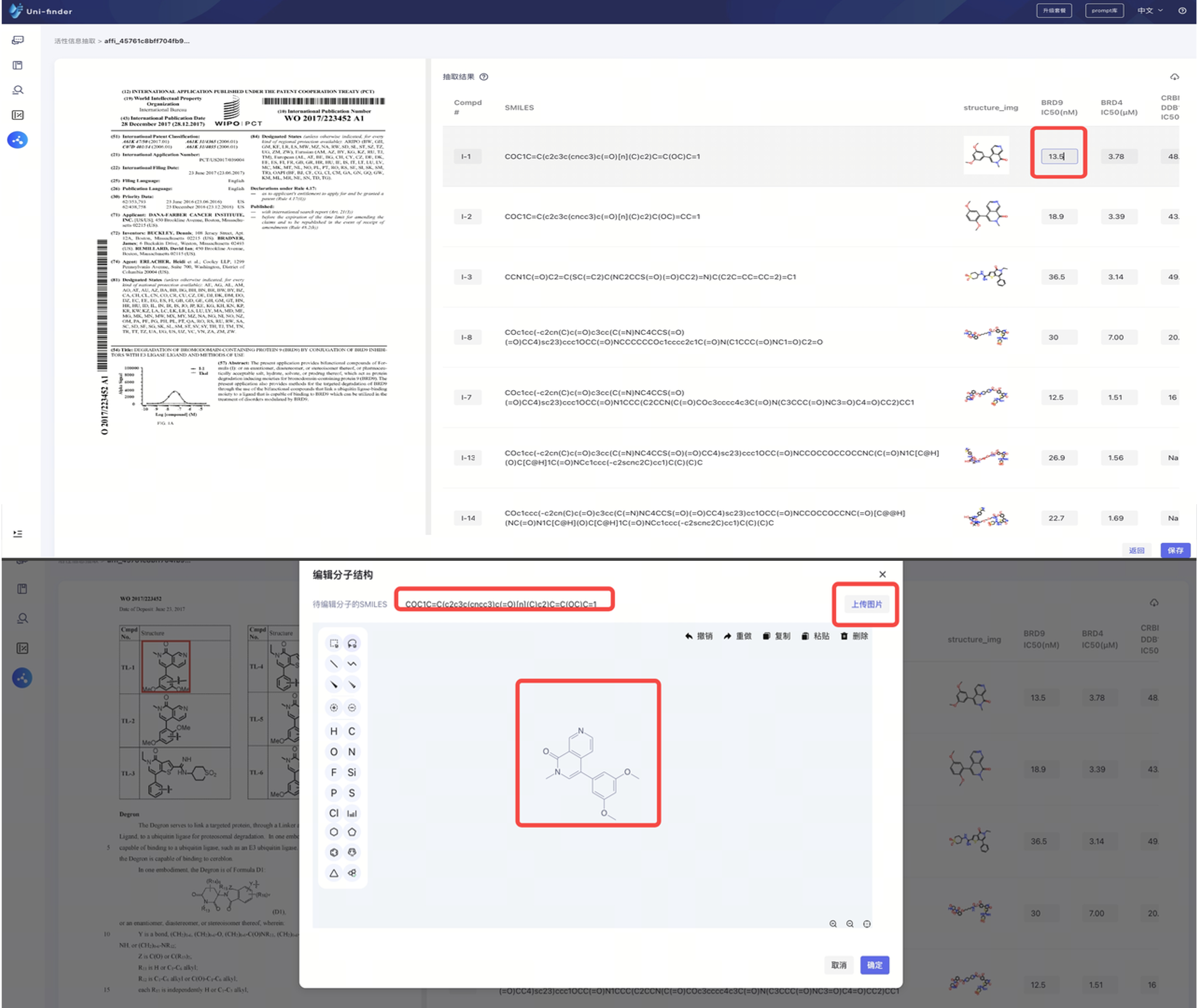}
        \caption{}
    \end{subfigure}\hfill
    \begin{subfigure}{0.48\textwidth}
        \centering
        \includegraphics[width=0.98\linewidth]{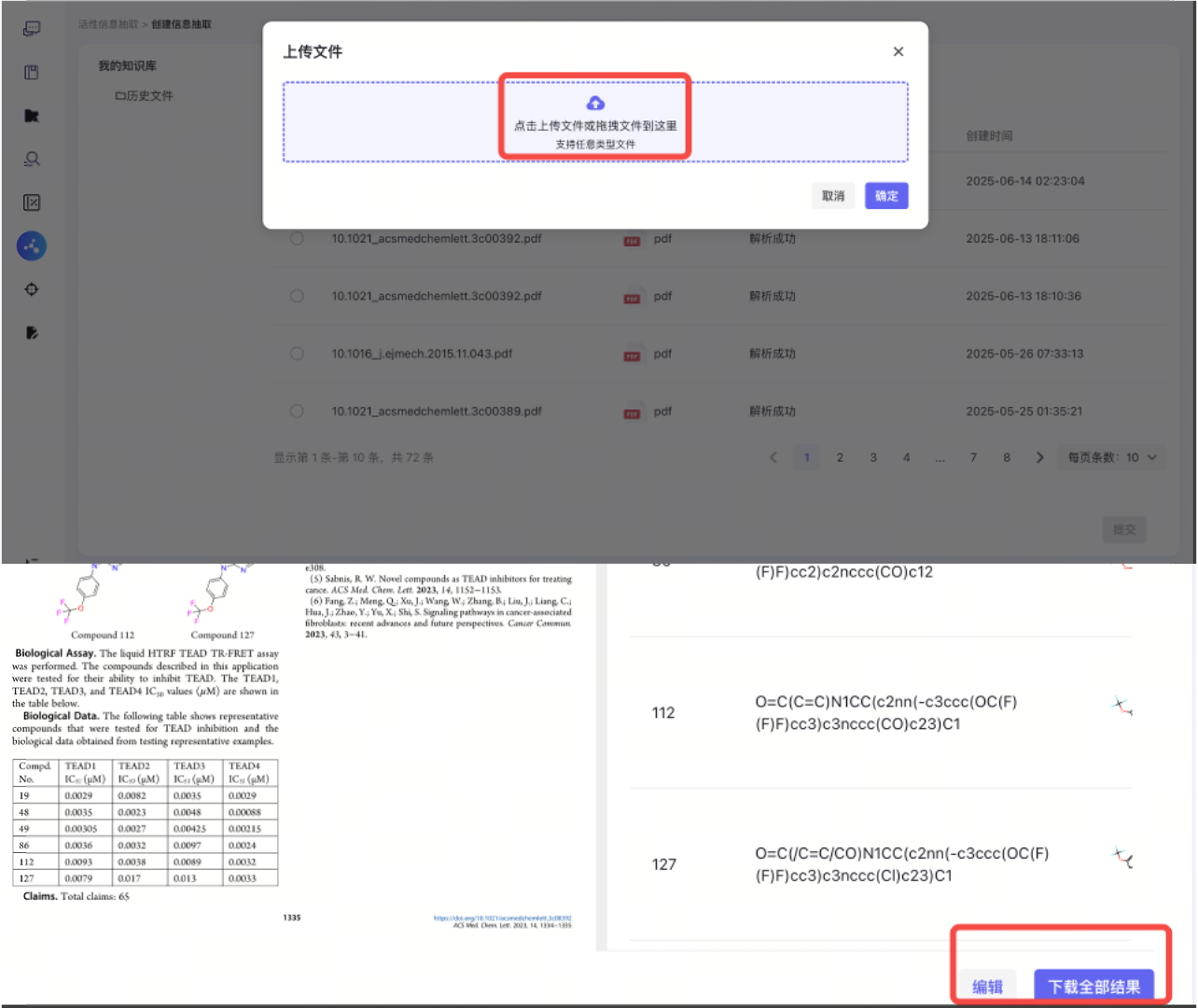}
        \caption{}
    \end{subfigure}

    \caption{Screenshots of the Doc2SAR web application: (a) Tabular display of extracted molecular activity data. (b) Example of traceability, highlighting the source of an extracted SMILES string in the original PDF. (c) Interface for interactive editing and correction of extracted results. (d)  Functionality for batch uploading files and downloading the extracted results.}
    \label{fig:webapp_results}
\end{figure*}

Upon completion of a task, the extracted information is presented in a clear, tabular format, as illustrated in Figure~\ref{fig:webapp_results}(a). Each row typically corresponds to an extracted molecule-activity pair, displaying the molecule coreference, its SMILES representation (potentially with a rendered 2D chemical structure image), and the associated activity values (e.g., IC50, EC50, units). A key feature of the web application is its traceability functionality. Users can click on an extracted SMILES string or an activity data point within the table, and the system will navigate to and highlight the corresponding source region in the original PDF document, as shown in Figure~\ref{fig:webapp_results}(b). This direct linking between the extracted data and its origin within the document is crucial for transparency, allowing for quick verification of the extraction accuracy and building user confidence.

Furthermore, recognizing that automated systems may occasionally require refinement, the web application incorporates an interactive editing interface (Figure~\ref{fig:webapp_results}(c)). Users can directly modify extracted coreferences, SMILES strings, or activity values. For SMILES correction, the interface supports direct text editing, drawing the chemical structure using an embedded molecular editor, or even uploading a corrected molecular image from which the SMILES can be re-derived. Finally, the curated and validated extraction results can be conveniently downloaded by the user in CSV format for downstream analysis and integration into their research workflows (Figure~\ref{fig:webapp_results}(d)). 
% A live demonstration of the web application is accessible at \textit{https://uni-finder.dp.tech/molecular-extract}.

% \begin{figure}[ht]
%     \centering
%     \begin{subfigure}{0.33\textwidth}
%         \centering
%         \includegraphics[width=0.97\linewidth]{figs/fig3a.png}
%         \caption{}
%     \end{subfigure}\hfill
%     \begin{subfigure}{0.33\textwidth}
%         \centering
%         \includegraphics[width=0.95\linewidth]{figs/fig3b.png}
%         \caption{}
%     \end{subfigure}\hfill
%     \begin{subfigure}{0.33\textwidth}
%         \centering
%         \includegraphics[width=0.95\linewidth]{figs/fig3c.png}
%         \caption{}
%     \end{subfigure}
%     \caption{Screenshots of the Doc2SAR web application: (a) Tabular display of extracted molecular activity data. (b) Example of traceability, highlighting the source of an extracted SMILES string in the original PDF. (c) Interface for interactive editing and correction of extracted results.}
%     \label{fig:webapp_results}
% \end{figure}

\clearpage
% \section{DocSAR-200 Data List}

\begin{table*}[t]
\centering
\caption{DocSAR-200 Data List}
{\scriptsize
\renewcommand{\arraystretch}{1.3}
\setlength{\tabcolsep}{4pt}
\begin{tabular}{c c c | c c c}
\toprule
\multicolumn{3}{c}{\textbf{Literature}} & \multicolumn{3}{c}{\textbf{Patent}} \\
\midrule
10.1021\_acsmedchemlett.4c00408 & 10.1021\_acsmedchemlett.3c00180 & 10.1016\_j.ejmech.2015.11.043 & BR112016001107B1 & US20160355824A1 & CN102482273B \\
10.1021\_acsmedchemlett.3c00475 & 10.1021\_jm025553u & 10.1021\_acsmedchemlett.4c00133 & US11155555 & CA3012408C & ES2988680T3 \\
10.1016\_j.bmcl.2017.03.030 & 10.1021\_acsmedchemlett.3c00222 & 10.1021\_acsmedchemlett.4c00261 & CA3114414A1 & AU2015238300C1 & US10100308 \\
10.1021\_acsmedchemlett.4c00230 & 10.1021\_acsmedchemlett.4c00168 & 10.1021\_jm050499d & CN102089304A & CN103402511B & US20110046151A1 \\
10.1021\_acsmedchemlett.3c00544 & 10.1016\_j.ejmech.2023.116113 & 10.1021\_jm049503w & JP2018519304A & CA2982708C & US10435414 \\
10.1021\_acsmedchemlett.4c00195 & 10.1021\_jm0600139 & 10.1021\_acsmedchemlett.4c00286 & CN102159555A & DK3122742T3 & US20130172386A1 \\
10.1021\_acsmedchemlett.3c00444 & 10.1021\_acsmedchemlett.3c00507 & 10.1021\_jm800936s & ES2675791T3 & WO2014113407A2 & CN102574816A \\
10.1016\_j.ejmech.2023.116098 & 10.1021\_jm990622z & 10.1021\_acsmedchemlett.3c00312 & TW201408641A & WO2008134474A2 & KR20230154230A \\
10.1021\_acsmedchemlett.4c00331 & 10.1021\_acsmedchemlett.4c00390 & 10.1021\_acsmedchemlett.4c00329 & ES2904513T3 & US20170298047A1 & BRPI0909017A2 \\
10.1021\_jm049660v & 10.1021\_acsmedchemlett.3c00392 & 10.1021\_jm0155042 & US10435407 & EA040922B1 & CN105793260B \\
10.1021\_jm970540f & 10.1021\_acsmedchemlett.4c00076 & 10.1002\_cmdc.202400302 & EP3408265B1 & DK2638031T3 & US20230303530A1 \\
10.1021\_jm050413g & 10.1021\_jm701228e & 10.1021\_acsmedchemlett.2c00406 & AU2022263269A1 & BR112016021507B1 & CA2479109C \\
10.1021\_jm801101z & 10.1021\_acsmedchemlett.3c00506 & 10.1016\_j.bmcl.2017.07.029 & KR102306071B1 & EP3464318B1 & CA2600052A1 \\
10.1021\_jm015573g & 10.1021\_acsmedchemlett.3c00308 & 10.1021\_jm060484v & US8598357 & AU2017216212B2 & US8703811 \\
10.1021\_acsmedchemlett.3c00494 & 10.1021\_jm990577v & 10.1021\_acsmedchemlett.3c00249 & ES2357988T3 & US20220056009A1 & CN1659133A \\
10.1021\_acsmedchemlett.4c00046 & 10.1021\_acsmedchemlett.4c00177 & 10.1021\_acsmedchemlett.4c00056 & EP2094694B1 & US20190209539A1 & US20240092754A1 \\
10.1016\_j.bmcl.2024.129931 & 10.1021\_acsmedchemlett.4c00217 & 10.1021\_acsmedchemlett.4c00137 & TW201024308A & EP3271333B1 & ES2577829T3 \\
10.1021\_acsmedchemlett.3c00393 & 10.1021\_acsmedchemlett.4c00285 & 10.1021\_acsmedchemlett.4c00414 & ES2626908T3 & AU2022230795A1 & JP2024540522A \\
10.1021\_acsmedchemlett.4c00119 & 10.1016\_j.bmcl.2024.130012 & 10.1021\_acsmedchemlett.3c00236 & EA031639B1 & US20230192660A1 & CN109153640B \\
10.1021\_acsmedchemlett.3c00234 & 10.1021\_acsmedchemlett.4c00303 & 10.1021\_jm0002679 & CN103080089A & CN107531683B & AU2017281903B2 \\
10.1016\_j.bmcl.2024.129725 & 10.1021\_jm049156q & 10.1021\_acsmedchemlett.4c00356 & US11883500 & US9006271 & US10406157 \\
10.1016\_j.bmcl.2005.05.013 & 10.1021\_acsmedchemlett.4c00332 & 10.1021\_acsmedchemlett.3c00564 & AU2012217150A1 & EP2256117A1 & AU2007321019B2 \\
10.1016\_j.bmcl.2017.06.049 & 10.1021\_acsmedchemlett.3c00389 & 10.1002\_cmdc.200900097 & JP6898933B2 & ES2991765T3 & US20170209432A1 \\
10.1021\_acsmedchemlett.3c00323 & 10.1021\_acsmedchemlett.4c00001 & 10.1021\_acsmedchemlett.3c00177 & EA035188B1 & CN108434147A & CN107849044A \\
10.1021\_acsmedchemlett.4c00118 & 10.1021\_acsmedchemlett.5c00128 & 10.1021\_acsmedchemlett.3c00441 & EP3552630A1 & CA3238252A1 & US20200255401A1 \\
10.1021\_acsmedchemlett.3c00271 & 10.1021\_acsmedchemlett.5c00107 & 10.1016\_s0223-5234(00)00142-2 & US20220056540A1 & US20190308987A1 & US9132145 \\
10.1021\_acsmedchemlett.4c00304 & 10.1021\_acsmedchemlett.3c00324 & 10.1016\_j.bmcl.2024.129965 & EP1716145B1 & US20180008587A1 & ES2546430T3 \\
10.1021\_acsmedchemlett.3c00235 & 10.1021\_acsmedchemlett.2c00395 & 10.1016\_j.bmcl.2012.04.036 & EP2091939B1 & EA046921B1 & US20190071444A1 \\
10.1021\_jm0256068 & 10.1021\_acsmedchemlett.4c00228 & 10.1021\_acsmedchemlett.4c00343 & JP2013530954A & CN108290866A & CA2806647C \\
10.1021\_acsmedchemlett.3c00554 & 10.1021\_jm980053f & 10.1021\_jm0610245 & CA2767372A1 & AU2016247476B2 & US20180208570A1 \\
10.1021\_acsmedchemlett.3c00477 & 10.1016\_j.bmcl.2018.07.049 & 10.1021\_acsmedchemlett.4c00011 & US20210380533A1 & AU2003212335B8 & JP6990657B2 \\
10.1021\_acsmedchemlett.4c00262 & 10.1021\_acsmedchemlett.3c00474 & 10.1021\_acsmedchemlett.3c00458 & JP7002335B2 & EP3455221B1 & US20230203011A1 \\
10.1002\_cmdc.202100398 & 10.1021\_acsmedchemlett.3c00473 & 10.1016\_j.bmcl.2025.130230 &  &  &  \\
\bottomrule
\end{tabular}
}
\label{tab:docsar200}
\end{table*}

\end{document}